\documentclass[accepted]{uai2023} 

\usepackage[american]{babel}

\usepackage{natbib} 
\usepackage{mathtools} 
\usepackage{booktabs} 
\usepackage{tikz} 
\usepackage{csquotes}
\usepackage{amsfonts}
\usepackage{amsmath, amsthm, amssymb}
\usepackage{float}
\usepackage{caption}
\usepackage{subcaption}
\usepackage{appendix}

\newtheorem{proposition}{Proposition}

\newcommand{\N}{\mathbb{N}}
\newcommand{\E}{\mathbb{E}}

\newcommand{\del}{\operatorname{d}}

\newcommand{\pgivenx}{\prob( \cdot \given \vec{x})}
\newcommand{\hatpgivenx}{\hat \prob( \cdot \given \vec{x})}
\newcommand{\ymarg}{p(y)}
\newcommand{\ygivent}{p(y \given \vtheta)}
\newcommand{\yrv}{p(Y)}

\newcommand{\yrvgivent}{p(Y | \Theta)}
\newcommand{\levelone}{\mathbb{P}(\cY)}

\newcommand{\ksimplex}{\Delta_K}
\newcommand{\ksimplextwo}[1][K]{\Delta_{#1}^{(2)}}

\newcommand{\hyptwo}{F}
\newcommand{\ber}[1]{\mathcal{B}(#1)}
\newcommand{\ent}{H}
\newcommand{\mi}{I}
\newcommand{\dkl}{D_{\text{KL}}}
\newcommand{\half}{\tfrac{1}{2}}
\newcommand{\dirac}[1][1/2]{\delta_{#1}}
\newcommand{\unif}[1][0, 1]{\mathcal{U}[#1]}

\renewcommand{\vec}[1]{\boldsymbol{#1}}
\newcommand{\given}{\, | \,}

\newcommand{\vtheta}{\vec{\theta}}

\newcommand{\vthetatrue}{\vtheta^\ast}
\newcommand{\fromto}{\longrightarrow}

\newcommand*{\defeq}{\mathrel{\vcenter{\baselineskip0.5ex \lineskiplimit0pt
			\hbox{\footnotesize.}\hbox{\footnotesize.}}}%
=}

\newcommand{\cX}{\mathcal{X}}
\newcommand{\cY}{\mathcal{Y}}
\newcommand{\cH}{\mathcal{H}}

\newcommand{\prob}{p}

\title{Quantifying Aleatoric and Epistemic Uncertainty in Machine Learning: \\Are Conditional Entropy and Mutual Information Appropriate Measures?}

\author[1,3]{\href{mailto:<lisa.wimmer@stat.uni-muenchen.de>?Subject=Your UAI 2023 paper}{Lisa~Wimmer}{}}
\author[2,3]{Yusuf~Sale}
\author[2,3]{Paul~Hofman}
\author[1,3]{Bernd~Bischl}
\author[2,3]{Eyke~H\"ullermeier}
\affil[1]{%
    Department of Statistics\\
    LMU Munich\\
    Germany
}
\affil[2]{%
    Institute of Informatics\\
    LMU Munich\\
    Germany
}
\affil[3]{%
    Munich Center for Machine Learning (MCML)\\
    Germany
}  

\begin{document}

\maketitle

\begin{abstract}
The quantification of aleatoric and epistemic uncertainty in terms of conditional entropy and mutual information, respectively, has recently become quite common in machine learning. 
While the properties of these measures, which are rooted in information theory, seem appealing at first glance, we identify various incoherencies that call their appropriateness into question.
In addition to the measures themselves, we critically discuss the idea of an additive decomposition of total uncertainty into its aleatoric and epistemic constituents.
Experiments across different computer vision tasks support our theoretical findings and raise concerns about current practice in uncertainty quantification. 
\end{abstract}

\section{Introduction}\label{intro}

Estimating, analyzing, and handling uncertainty has always been an integral part of classical statistics.
More recently, the broader machine learning community has come to acknowledge its vital role in producing reliable estimates, which is, e.g., considered paramount in safety-critical applications \citep{senge_2014_ReliableClassificationLearning, kendall_2017_WhatUncertaintiesWe, psaros_2022_UncertaintyQuantificationScientificb}.
In the context of supervised learning, the focus is typically on \emph{predictive uncertainty}, i.e., the learner's uncertainty about the outcome $y \in \cY$  given a query instance $\vec{x}$ from an input space $\cX$.
It is often expedient to decompose the \emph{total} amount of uncertainty (TU) into its \emph{aleatoric} (AU) and \emph{epistemic} (EU) parts \citep{hullermeier_2021_AleatoricEpistemicUncertainty}.
The aleatoric component of TU arises from the stochastic nature of the mapping from $\cX$ to $\cY$, which implies that the \enquote{ground truth} is a (generally non-Dirac) conditional probability distribution $\pgivenx$ over $\cY$. 
Potential sources of AU include measurement errors, randomness in physical quantities, and the simple fact that $\vec{x}$ may not suffice to explain $y$ \citep{malinin_2019_UncertaintyEstimationDeep}. 
Thus, it is not possible to predict the outcome with certainty even assuming perfect knowledge of the data-generating process. 
In practice, the learner only finds an empirical estimate $\hatpgivenx$ from a pre-defined hypothesis space $\cH$.
Roughly speaking, EU then refers to a lack of knowledge about
the discrepancy between $p$ and $\hat{p}$ that can -- in contrast to AU -- be alleviated by gathering more training samples\footnote{ 
An additional source of EU is model misspecification, meaning that $p \not\in \cH$ (e.g., \citet{wilson_2020_CaseBayesianDeep}); although important, it is difficult to tackle formally and often omitted from the analysis.} \citep{senge_2014_ReliableClassificationLearning}.
Not least due to this potential reduction, the quantification of EU is helpful in many  applications of AI, e.g., in order to invoke human intervention in multi-stage decision processes \citep{gal_2017_DeepBayesianActivea, budd_2021_SurveyActiveLearning,  kirsch_2022_MarginalJointCrossEntropies}.

A specific set of measures for TU, AU, and EU has found broad consensus in the uncertainty quantification literature.
Promoted by \citet{houlsby_2011_BayesianActiveLearning} and \citet{depeweg_2019_ModelingEpistemicAleatoric}, among others, the core idea revolves around \emph{Shannon entropy} \citep{shannon_1948_MathematicalTheoryCommunication} as a measure for TU, which can be shown to (additively) decompose into \emph{conditional entropy}, interpreted as a measure of AU, and \emph{mutual information}, resulting as a measure of EU \citep{kendall_2017_WhatUncertaintiesWe, smith_2018_UnderstandingMeasuresUncertaintya, charpentier_2022_DisentanglingEpistemicAleatoric}.
Various applications have recently used these measures (e.g., \citet{michelmore_2018_EvaluatingUncertaintyQuantification, shelmanov_2021_ActiveLearningSequence, winkler_2022_StochasticControlBayesian}).

While the underlying ideas are well-established and mathematically well-founded, they originate from information theory and are not necessarily suited to uncertainty quantification in statistical learning. 
In fact, we find that the proposed entropy-related measures exhibit questionable behavior in a number of settings.
We thus argue for a critical re-assessment of the established practice.  
Our contributions are as follows: We (i) point out numerous incoherencies of uncertainty measures based on decomposing Shannon entropy, (ii) shed light on why an additive disaggregation of TU might not be meaningful, and (iii) provide evidence for the relevance of these observations in practical applications.

\section{Background}
\subsection{Uncertainty Representations}
\label{sub:representation}

In the following, we assume categorical target variables from a finite label space $\cY = \{ y_1, \ldots, y_K \}, K \in \N$. 
The set $\levelone$ of (first-order) probability distributions over $\cY$ can be identified with the ($K$-1)-simplex
$\ksimplex \defeq \left \{ \vtheta = (\theta_1, \ldots , \theta_K) \in [0,1]^K ~ \given~ \| \vtheta \|_1 = 1 \right \}$.
For each $\vtheta \in \ksimplex$, we can compute an associated degree of AU \citep{hullermeier_2021_AleatoricEpistemicUncertainty}.
To capture EU, the learner must be able to express uncertainty about $\vtheta$, which can be accomplished by a (second-order) probability distribution over (first-order) distributions $\vtheta$. 

The canonical approach to such a bi-level representation is Bayesian inference, where uncertainty about the (Bayes) predictor is translated into uncertainty about a prediction expressed in terms of the posterior predictive distribution \citep{gelman_2021_BayesianDataAnalysis}.
Another idea is to estimate EU directly along with the prediction of the outcome.
For instance, it is possible to learn the parameters of a Dirichlet distribution with a concentration parameter expressing EU 
\citep{malinin_2018_PredictiveUncertaintyEstimation, charpentier_2020_PosteriorNetworkUncertainty, huseljic_2020_SeparationAleatoricEpistemic, kopetzki_2021_EvaluatingRobustnessPredictive}.
Either way, we arrive at a second-order or second-order predictor $\hyptwo: \cX \fromto \ksimplextwo,$ where $\ksimplex^{(2)} = \mathbb{P} ( \mathbb{P} ( \cY) )$ denotes the set of second-order distributions over $\mathcal{Y}$. 
For the sake of simplicity, we subsequently omit the conditioning on the query instance $\vec{x}$.

Each probability vector $\vtheta \in \ksimplex$ implies a categorical distribution on $\cY$, such that $\theta_k = \prob(y_k \given \vtheta)$ is the probability of outcome $y_k$, and one of these distributions, $\vthetatrue$, corresponds to the (unknown) ground truth. 
If $\hyptwo(\vec{x}) = Q \in \ksimplextwo$ is the prediction for a query $\vec{x}$, then $Q$ assigns a probability $Q(\vtheta)$ to each distribution $\vtheta \in \ksimplex$. 
$Q$ represents the epistemic state of the learner about $\vec{x}$, and a stronger concentration of $Q$ indicates more certainty about the true distribution $\vthetatrue$.
    Note that we immediately regard the output distribution and omit from our notation any relation to objects that play a role in estimating $\vtheta$ (e.g., the Bayes posterior is often expressed as a distribution over model parameters given data).

\begin{figure}[H]
    \begin{center}
        \includegraphics[width=0.4\textwidth]{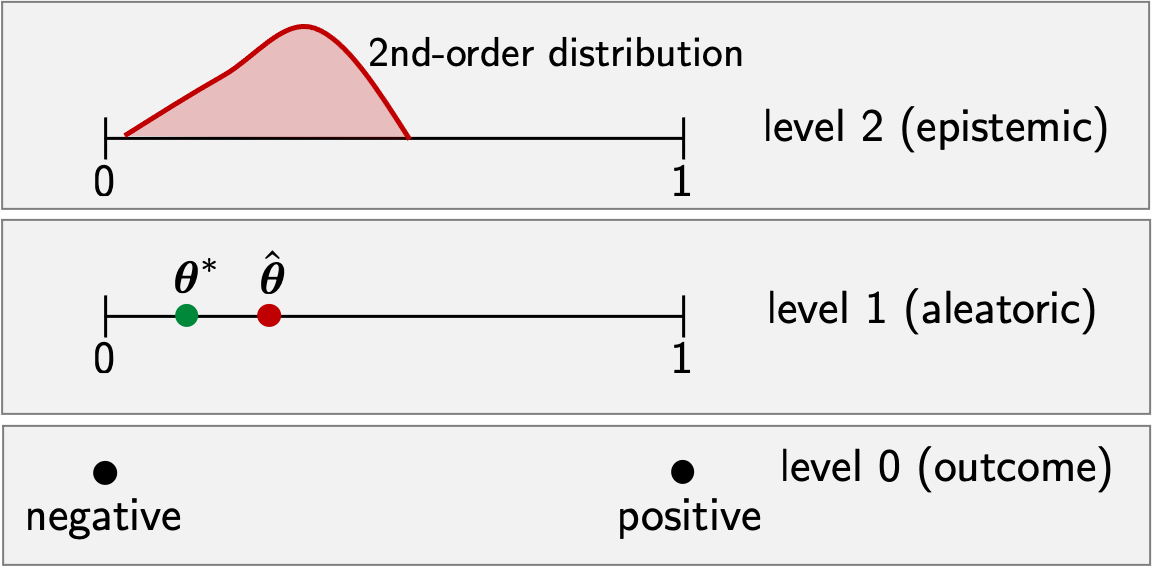}
        \caption{
        Bi-level uncertainty representation in binary classification.
        We pose a second-order distribution $Q$ over Bernoulli distributions corresponding to the probability $\theta$ of the positive class.
        In Bayesian inference, $Q$ is given by the posterior distribution (epistemic level), and a point prediction $\hat{\theta}$ (aleatoric level) is obtained by model averaging.
        On level zero, only binary class labels are observed.
        }
        \label{fig:bayespp}
    \end{center}
\end{figure}

As an example, consider the case of binary classification ($K=2$), where $\cY = \{ y_0, y_1\}$ consists of only two classes, negative and positive (illustrated in Fig.~\ref{fig:bayespp}). 
The conditional distribution $\pgivenx$ is now determined by the probability vector $\vtheta = (\theta_0 , \theta_1) = (\prob( y_0 \given \vec{x}), ~\prob( y_1 \given \vec{x}))$. 
As the individual class probabilities sum up to one, prediction in $\vec{x}$ effectively comes down to inference about the ground-truth probability $\theta_1^\ast = \prob( y_1 \given \vec{x})$ of the positive class, and hence about the success parameter $\theta$ of a Bernoulli distribution $\ber{\theta}$. The corresponding second-order distribution $Q$ could be, for example, a Beta distribution (or, more generally, a Dirichlet distribution for $K \geq 2$).

\subsection{Entropy-Based Information Measures}
\label{sub:infometrics}

Given a representation of an uncertain prediction in terms of a second-order distribution $Q$, one is often interested in quantifying the learner's uncertainty in terms of a single number, and decomposing the TU into its aleatoric and epistemic contributions. 
One approach that is now commonly adopted \citep{houlsby_2011_BayesianActiveLearning, gal_2016_UncertaintyDeepLearning, depeweg_2018_DecompositionUncertaintyBayesian, smith_2018_UnderstandingMeasuresUncertaintya, mobiny_2021_DropConnectEffectiveModeling} is based on information-theoretic quantities derived from Shannon entropy.
Let $Y$ signify the random outcome variable with marginal probability 
\begin{equation}
    \ymarg = \textstyle \int_{\ksimplex} \ygivent \, \del Q(\vtheta).
\end{equation}
The (discrete) Shannon entropy of $Y$ can be defined as
\begin{align} \label{eq:shannon}
    \ent(Y) 
    = - \textstyle \sum_{y \in \cY} \ymarg \cdot \log \ymarg \, , 
\end{align}
where the logarithm is typically set to base 2.
For continuous label spaces, we obtain the \emph{differential} entropy analogously by replacing the sum in Eq.~(\ref{eq:shannon}) with an integral over $\cY$ \citep{cover_2006_ElementsInformationTheory}.

Shannon entropy can be interpreted as the degree of uniformity in the distribution of a random variable, or the amount of information to be gained from observing its realization (further analogies exist in physics and source coding).
It enjoys a number of desirable properties: it is non-negative, maximal for the uniform distribution, and invariant to permutations of classes in $\cY$ \citep{cover_2006_ElementsInformationTheory}.

Exploiting the fact that $Y$ is the expectation of the conditional outcome (given $\vtheta$) with respect to the second-order distribution $Q$, the Shannon entropy, as a measure of TU, can be computed as
\begin{align}
    \ent(Y) &= \ent \big( \E_{Q} \left[ Y \given \vtheta \right] \big). \label{eq:ent}
\end{align}
A trivial result from information theory states that Shannon entropy additively decomposes into \emph{conditional entropy} and \emph{mutual information} (e.g., \citet{ash_1965_InformationTheory}):
\begin{equation}\label{eq:dec}
    \ent(Y) = \ent(Y \given \Theta) + \mi(Y, \Theta).
\end{equation}
Here, $\Theta$ denotes the random variable of first-order distributions $\vtheta \sim Q$.
Conditional entropy $\ent(Y \given \Theta)$ expresses the uncertainty about $Y$ that would remain if the realization of $\Theta$ were known, and is therefore taken to quantify AU.
The motivation for expressing AU this way is as follows: By fixing a first-order distribution $\vtheta \in \ksimplex$, all EU is essentially removed and only AU remains, such that $\ent (Y \given \vtheta) = - \sum_{k=1}^K \theta_k \, \log \theta_k$ is indeed a natural measure of AU. 
However, as $\vtheta$ is not precisely known, we take the expectation with respect to the second-order distribution: 
\begin{align}
    \ent (Y \given \Theta) &= \E_{Q} \left[ \ent \left( Y \given \vtheta\right) \right] \label{eq:condent} \\ \notag
    &= \textstyle \int \limits_{\ksimplex} \underbrace{ \textstyle \sum_{y \in \cY} - \prob(y \given \vtheta) \log \prob(y \given \vtheta) }_{H(Y \given \vtheta)} \, \del Q( \vtheta ) .
\end{align}
Mutual information $\mi(Y, \Theta)$, proposed to embody EU, is a symmetric measure for the expected information gained about one variable through observing the other and vanishes if both variables are independent.
Intuitively, it quantifies the potential reduction in uncertainty about $Y$ through observing $\Theta$\footnote{
    It should be noted that \citet{houlsby_2011_BayesianActiveLearning} proposed the entropy decomposition to inform uncertainty sampling in active learning, where mutual information has a slightly different purpose, namely reducing uncertainty about $\Theta$, while we use $\mi(Y, \Theta)$ to quantify uncertainty about $Y$ (which the symmetric nature of $\mi(\cdot)$ enables us to do).
} \citep{ash_1965_InformationTheory}:
\begin{align} \notag
    \mi(Y, \Theta) &= \ent(Y) - \ent(Y \given \Theta) \label{eq:mi_res} \\ 
    &= \E_{Q} \left[ \dkl \left( \yrvgivent ~\|~ \yrv \right) \right].
\end{align}

\subsection{Finite-Ensemble Approximation}
\label{sub:ensemble}

In practice, it will hardly be possible to compute exact expectations with respect to $Q$ due to the intractability of the corresponding integrals, so we must permit some kind of approximation.
The standard approach to this problem is Monte Carlo integration with a finite number $M \in \N$ of samples $\vtheta \sim Q$, whose induced mean, by the law of large numbers, almost surely converges to the true integral \citep{andrieu_2003_IntroductionMCMCMachine}.
In machine learning, this way of combining the predictions of multiple models is common practice and known as \emph{ensemble learning} (e.g., \citet{breiman_2001_RandomForests, lakshminarayanan_2017_SimpleScalablePredictive}).
For TU, the entropy over the approximate first-order distribution becomes
\begin{equation} \label{eq:emptu}
    \ent(Y) =  \ent \left( \tfrac{1}{M} \textstyle \sum_{m=1}^M p \left(y \given \vtheta^{(m)} \right) \right),
\end{equation}
where $\vtheta^{(m)}$, $m \in \{1, \ldots , M\}$, is the prediction of the $m$-th ensemble member.
Analogously, we obtain the ensemble conditional entropy as a measure of AU:
\begin{equation} \label{eq:empau}
    \ent( Y \given \Theta ) = \tfrac{1}{M} \textstyle \sum_{m=1}^M \ent \left(p \left(y \given \vtheta^{(m)} \right) \right).
\end{equation}
The discrete EU is the Jensen-Shannon divergence
\begin{equation}\label{eq:js}
\tfrac{1}{M} \textstyle \sum_{m=1}^M \dkl \left( p \left(y \given \vtheta^{(m)} \right) \, \Big\Vert \, \tfrac{1}{M} \textstyle \sum_{j=1}^M p \left(y \given \vtheta^{(j)} \right)  \right). 
\end{equation}
Finite-ensemble EU still results as the difference between Eq.~(\ref{eq:emptu}) and Eq.~(\ref{eq:empau}) under the conditions of Section~\ref{sub:infometrics}.
Obviously, ensemble learning provides a rather coarse approximation to the expectation $\E_{Q} (\cdot)$, especially if the size $M$ of the ensemble is small.
Consider, for example, deep ensembles \citep{lakshminarayanan_2017_SimpleScalablePredictive}, which have become a \emph{de facto} gold standard in probabilistic machine learning \citep{ovadia_can_2019, ashukha_2020_PitfallsInDomainUncertaintya, psaros_2022_UncertaintyQuantificationScientificb}.
The cost of training each of their neural network members is typically so high that $M$ remains in the realm of five to ten (e.g., \citet{abe_2022_DeepEnsemblesWork}, \citet{turkoglu_2022_FiLMEnsembleProbabilisticDeep}). We must then expect certain (low-density) regions in the space of first-order distributions to be systematically undersampled, affecting the estimate of EU in particular.

\section{Related Work}

The decomposition in Eq.~(\ref{eq:dec}) is widely applied and often used in conjunction with Bayesian learning, e.g., to obtain more robust predictions by explicitly modeling both components of TU \citep{kendall_2017_WhatUncertaintiesWe}, to design better optimization procedures by using the decomposition in stochastic differential equations \citep{winkler_2022_StochasticControlBayesian}, or to improve the general understanding of neural networks \citep{woo_2022_AnalyticMutualInformation}.
In natural language processing, \citet{wu_2020_EnsembleApproachesUncertainty} filter unreliable predictions in spoken language assessment caused by high EU, while \citet{shelmanov_2021_ActiveLearningSequence} perform active learning with EU-based acquisition for label-efficient sequence tagging.
Further fields of application include autonomous driving, where EU is found to be predictive of forthcoming accidents \citep{michelmore_2018_EvaluatingUncertaintyQuantification}, and medical diagnostics, with a focus on identifying inherently ambiguous cases  \citep{mobiny_2021_DropConnectEffectiveModeling}.

We find, however, that the entropy decomposition may lead to unintended results discussed in the subsequent section.
While we do not dispute the mathematical correctness of Eq.~(\ref{eq:dec})-(\ref{eq:mi_res}), we argue that the individual quantities and their additive aggregation are not suitable for evaluating predictive uncertainty.
Although previous work uncovered similar shortcomings of measures in related uncertainty frameworks (e.g., \citet{pal_1992_UncertaintyMeasuresEvidential} on evidential reasoning), the entropy-based measures -- to the best of our knowledge -- have drawn hardly any criticism in the machine learning community.
In the following, we point out numerous inconsistencies that cast doubt on the usefulness of the individual measures and address the more general issue of additive decomposition.

\section{Critical Assessment}
\label{sec:critical}

Let TU, AU, and EU denote, respectively, measures $\ksimplextwo \fromto \mathbb{R}$ of total, aleatoric, and epistemic uncertainty associated with a (second-order) uncertainty representation $Q \in \ksimplextwo$. In the literature, it is common to define the fundamental properties of uncertainty measures through an axiomatic approach (see also, for example, \citet{bronevich2008axioms, pal1993uncertainty}. In the following, we define some properties these measures should naturally satisfy.
\begin{itemize}\itemsep0mm
\item[A0] TU, AU, and EU are non-negative. 
\item[A1] EU vanishes for Dirac measures $Q = \delta_{\vec{\theta}}$.
\item[A2] EU and TU are maximal for $Q$ being the uniform distribution on $\ksimplex^{(2)}$.
\item[A3] If $Q^\prime$ is a mean-preserving spread\footnote{Let $X \sim Q, X^\prime \sim Q^\prime$ be two random variables, where $Q, Q^\prime \in \ksimplextwo$. Then, $Q^\prime$ is called a mean-preserving spread of $Q$ iff $X^\prime \overset{d}{=} X + Z$, for some random variable $Z$ with $\mathbb{E}[Z| X = x] = 0 ~ \forall x$ in the support of $X$.} of $Q$, then $\text{EU}(Q^\prime) \geq \text{EU}(Q)$ (weak version) or $\text{EU}(Q^\prime) > \text{EU}(Q)$ (strict version); the same holds for TU. 
\item[A4] If $Q^\prime$ is a center-shift\footnote{
$Q$ and $Q^\prime$ differ only in their respective means, where the mean of $Q^\prime$ is closer
    -- in terms of $L1$ distance in Cartesian coordinates -- 
to the barycenter of $\ksimplextwo$.
} of $Q$, then $\text{AU}(Q^\prime) \geq \text{AU}(Q)$ (weak version) or $\text{AU}(Q^\prime) > \text{AU}(Q)$ (strict version); the same holds for TU.
    \item[A5] If $Q^\prime$ is a spread-preserving location shift\footnote{
    $Q$ and $Q^\prime$ differ only in their respective means.
    } of $Q$, then $\text{EU}(Q^\prime) = \text{EU}(Q)$. 
\end{itemize}

\subsection{Total Uncertainty}

\begin{figure}[H]
    \begin{center}
        \centering
        \includegraphics[width=0.45\textwidth]{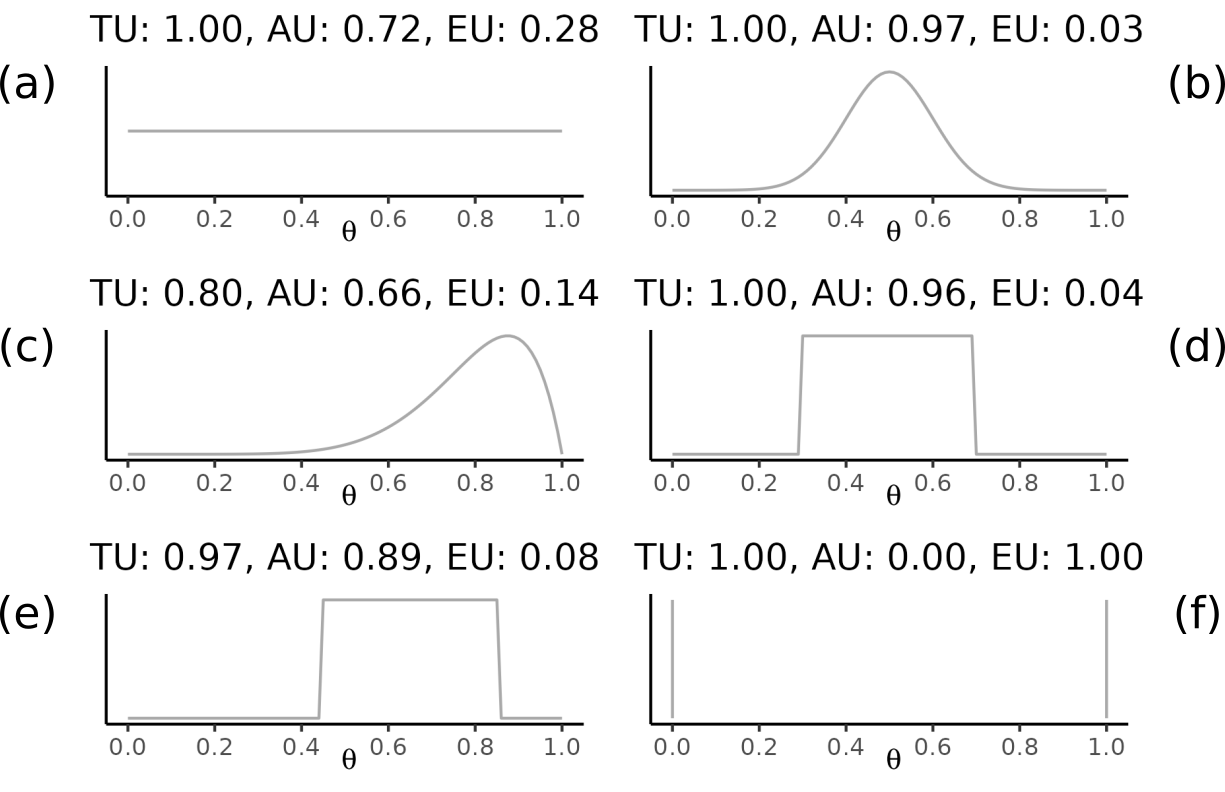}
        \caption{
        Different second-order distributions $Q$ over the parameter $\theta$ of a Bernoulli distribution with associated uncertainty.
                    (a) $\unif$; (b) $\mathcal{N}(0.5, 0.01)$; (c) $\operatorname{Beta}(8, 2)$; (d) $\unif[0.3, 0.7]$; (e) $\unif[0.45, 0.85]$; (f) $\half \dirac[0] + \half \dirac[1]$.
        }
        \label{fig:jensen}
    \end{center}
\end{figure}

Although Shannon entropy is widely established as a measure of uncertainty associated with a random variable, one may wonder about its appropriateness for quantifying total predictive uncertainty in the scenario outlined above.
As an illustration, we pick up our previous example of binary classification (note that properties A0-A5 apply to $Q, Q^\prime \in \ksimplextwo$ but violations for $K=2$ suffice to conclude that they do not hold in general).
Fig.~\ref{fig:jensen} shows some exemplary second-order distributions $Q$ together with the values they induce for TU, AU, and EU about the prediction of $\theta$.

We first note that property A3 is violated by TU in terms of entropy, at least in its strict version, because 
 (\ref{eq:ent}) depends on $Q$ only through its expectation. 
\begin{proposition}
\label{prop:41}
Total uncertainty defined in terms of Shannon entropy (\ref{eq:ent}) violates the strict version of A3. 
\end{proposition}
For example, in the special case $Q \in \ksimplextwo[2]$, $H(Y)$ is maximal as soon as $Q$ is symmetric around $\theta = \half$. Consequently, the uniform distribution on the unit interval, $Q = \unif$ (Fig.~\ref{fig:jensen}a), which is commonly considered a representation of \emph{complete ignorance} (justified by the \enquote{principle of indifference} invoked by Laplace, or by referring to the principle of maximum entropy), has the same TU as a mixture of Dirac distributions $Q= \half \dirac[0] + \half \dirac[1]$ (Fig.~\ref{fig:jensen}f) dividing all probability mass between the outcomes $\theta = 0$ and $\theta = 1$. 
While one may argue that the uncertainty about the target variable $Y$ is indeed the same in both cases, it seems counter-intuitive to obtain identical results when in one scenario, the learner (supposedly) knows nothing, whereas, in the other, it has substantial knowledge about the ground-truth distribution $\theta^\ast$.
Likewise, it appears strange that any distribution $Q$ not symmetric around $\half$ will have a lower TU, including, for example, $\unif[0.45, 0.85]$ (Fig.~\ref{fig:jensen}e), when it arguably represents a lower degree of informedness than the Dirac mixture.

The root cause of these peculiarities is the role of the distribution $Q$ in the computation of TU -- effectively, $Q$ is marginalized over. 
In this sense, the reason for having maximal TU in the case of $\unif$ is not that it (supposedly) encodes the greatest lack of knowledge, but only its symmetry around $\theta = \half$. 
Information about the second-order distribution is thus compressed to a single number in the form of its expectation.
On the other hand, the concentration of $Q$ is hardly reflected in the TU, although it should be an important indicator of the learner's state of knowledge.

\subsection{Aleatoric Uncertainty}

    Noting that, in Eq.~(\ref{eq:condent}), $Q$ represents the learner's \emph{subjective belief} about a supposedly objective ground-truth $\vthetatrue$ associated with query $\vec{x}$, $\ent (Y \given \Theta)$ can be seen as the learner's (current) best guess of the true AU. 
    Although this might seem a sensible quantity, the relationship between this estimate and the true AU, $\ent(Y \given \vthetatrue)$, is by no means clear. 
    The problem is that AU estimates -- even if the additive nature of the decomposition might suggest otherwise -- cannot be separated from the learner's epistemic state.
    In the finite-sample case, the learner must fall short of perfect knowledge, meaning that its belief $Q$ will not coincide with the Dirac distribution on $\vthetatrue$ that represents the true second-order distribution.
    This, however, directly translates to the estimate of AU that is derived by taking the expectation over $Q$.
    Whenever there is EU, we must expect the AU estimation to be compromised, and the absence of a ground-truth precludes us from knowing how it is affected exactly. 

In particular, conditional entropy is neither a true expectation nor a lower or upper bound on the true AU, and the resulting point estimate suggests a degree of informedness that is hardly justified.
The limited ability of probability distributions to properly distinguish between different subjective beliefs adds to the problem.
For example, if a uniform $Q$ is meant to represent complete ignorance, then all the learner can infer is that the true AU is between zero (for $\theta \in \{0, 1\}$) and one (for $\theta = \half$). 
This is clearly a different situation than having perfect knowledge about all first-order distributions being equally probable.
Ultimately, however, both lead to the same average over the support of $Q$.
Therefore, we must conclude that AU is computed with respect to a belief that the second-order distribution may fail to express, and that must be expected to be wrong.

\subsection{Epistemic Uncertainty}

The definition of mutual information (Eq.~(\ref{eq:mi_res})) may appear intuitively plausible at first sight: If the observation of $Y$ reveals a sizeable amount of information about $\Theta$, then the learner's uncertainty about $\Theta$ must be high (\textit{et vice versa}). 
Likewise, a high divergence in the finite-ensemble version (Eq.~(\ref{eq:js})), i.e., in the predictions of different ensemble members, 
can be viewed as indicating a high EU. 

That said, the measures do not necessarily behave as one may expect. 
Looking at Fig.~\ref{fig:jensen}, a first observation is that EU tends to be relatively low, which is especially noticeable in the case of the standard uniform distribution (upper left).
As already mentioned, $\unif$ is usually meant to represent complete ignorance, suggesting that the learner is as uninformed as it could be, so we would expect a measure of EU to fulfill property A2 (i.e., be maximal) in this case. 
\begin{proposition}
EU in terms of mutual information $I(Y, \Theta)$ violates property A2. 
\end{proposition}
This violation holds in general, and can be seen very easily for the case $K=2$ in Fig.~\ref{fig:jensen}. In Bayesian inference, for example, the uniform distribution is commonly adopted as a prior, representing knowledge before seeing any data. 
In such a state, one would assume all uncertainty to epistemic, and hence the measure of EU to be very high. 
On the contrary, however, AU (at 0.72) is well above EU (at 0.28). 
Actually, the maximal EU (1.00) is reached only for the case of Fig.~\ref{fig:jensen}f, in which $Q$ is a mixture of two Dirac measures, one placed at $\theta = 0$ and the other at $\theta= 1$. 
This distribution suggests a deterministic dependency, where the outcome is either certainly positive or certainly negative -- a case in which the learner undoubtedly knows more than nothing, so again, the level of EU does not seem appropriate.

Also, note that EU is not invariant against a location shift of the distribution $Q$, even though it would be reasonable to expect that EU remains constant when the spread of $Q$ does not change.
\begin{proposition}
There exist location shifts
$Q^{\prime} \in \ksimplextwo[K]$ of $Q \in \ksimplextwo[K]$ such that 
    property A5 is violated.
\end{proposition}
For example, the distributions in Fig.~\ref{fig:jensen}d and Fig.~\ref{fig:jensen}e (case $K=2$) are both uniform on an interval of length 0.4, suggesting the same level of informedness about the ground truth $\theta$. 
However, the respective degrees of EU (0.04 vs. 0.08) differ. 
Of course, one may argue that the uncertainty will depend not only on the shape but also on the location of the distribution. 
Then, however, $\unif[0.3, 0.7]$ should arguably exhibit higher uncertainty, as it puts more probability mass close to $\theta = \half$ than $\unif[0.45, 0.85]$.
This is indeed the case for TU and AU but exactly reversed for EU.

Lastly, we would expect EU to increase when $Q$'s mass is spread over a larger support (all else being equal), such that more hypotheses are deemed likely. 
Such spreading happens, for example, in moving from $\half \dirac[0.45] + \half \dirac[0.95]$ to $\unif[0, 1]$, while EU actually gets higher rather than lower.    

\begin{proposition}
EU in terms of mutual information $I(Y, \Theta)$ violates property A3. 
\end{proposition}

Mathematically, the numbers are clearly correct and meaningful: In Fig.~\ref{fig:jensen}f, for example, the ground-truth distribution $\theta$ is uniquely determined as soon as the outcome $Y$ has been observed, so mutual information should be maximal. 
What must be concluded is rather that mutual information may not be the right measure of EU. 
Indeed, mutual information is arguably more a measure of \emph{divergence} or \emph{conflict} than of ignorance, as the common understanding of EU would suggest.
This is quite obvious from Eq.~(\ref{eq:mi_res}) and even easier to see in the discrete version (Eq.~(\ref{eq:js})) for finite ensembles.
Mutual information effectively quantifies the expected divergence of single hypotheses from the opinion given by integrating over all of them.
Consequently, it is much higher for the Dirac mixture in Fig.~\ref{fig:jensen}f, with maximal divergence of both hypotheses from the average $\theta = \half$, than for $\unif$, which also assigns probability mass to many alternatives that diverge only little from the expectation.
Thus, mutual information behaves as expected but does not seem well-suited to measuring a quantity taken to represent \emph{ignorance}. 

\subsection{Additive Decomposition}
\label{sub:additive}

Let us conclude this section with a few remarks on another potential problem of uncertainty quantification, which is related, though not restricted, to the quantification discussed in this paper. 
According to Eq.~(\ref{eq:dec}), TU (entropy) decomposes \emph{additively} into AU (conditional entropy) and EU (mutual information). 
When combining distinct sources of uncertainty, additive representations of this kind appear natural and can also be found for other measures of uncertainty, such as variance (see, e.g., \citet{depeweg_2018_DecompositionUncertaintyBayesian} for a discussion of entropy and variance decomposition). 
However, when considering Eq.~(\ref{eq:dec}) in a machine learning setting, where the (total) uncertainty is not fixed but decreases with increasing sample size, additivity might be less obvious because the aleatoric part does not correspond to the \emph{true} AU, but only to an \emph{estimate} thereof. 
Moreover, this estimate depends on the EU, such that the two measures are tightly interwoven and certainly not independent of each other. 
So, what is the justification for simply adding them up?

\begin{figure}[H]
    \begin{center}
      \includegraphics[width=0.35\textwidth, trim=20 0 20 20, clip]{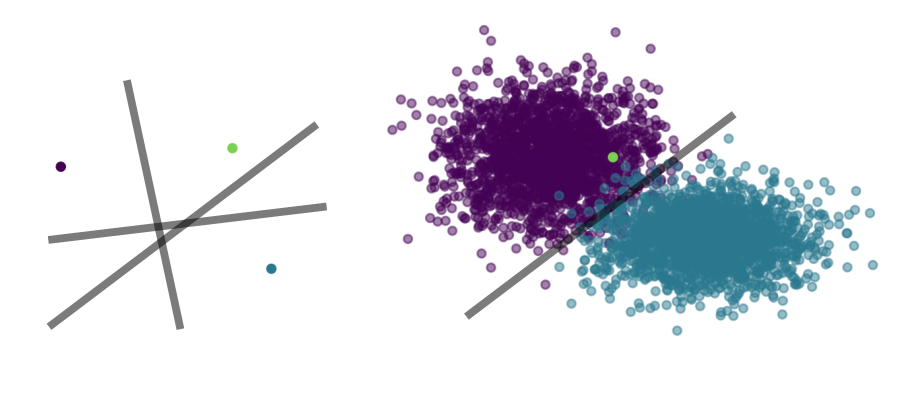} 
      \includegraphics[width=0.3\textwidth]{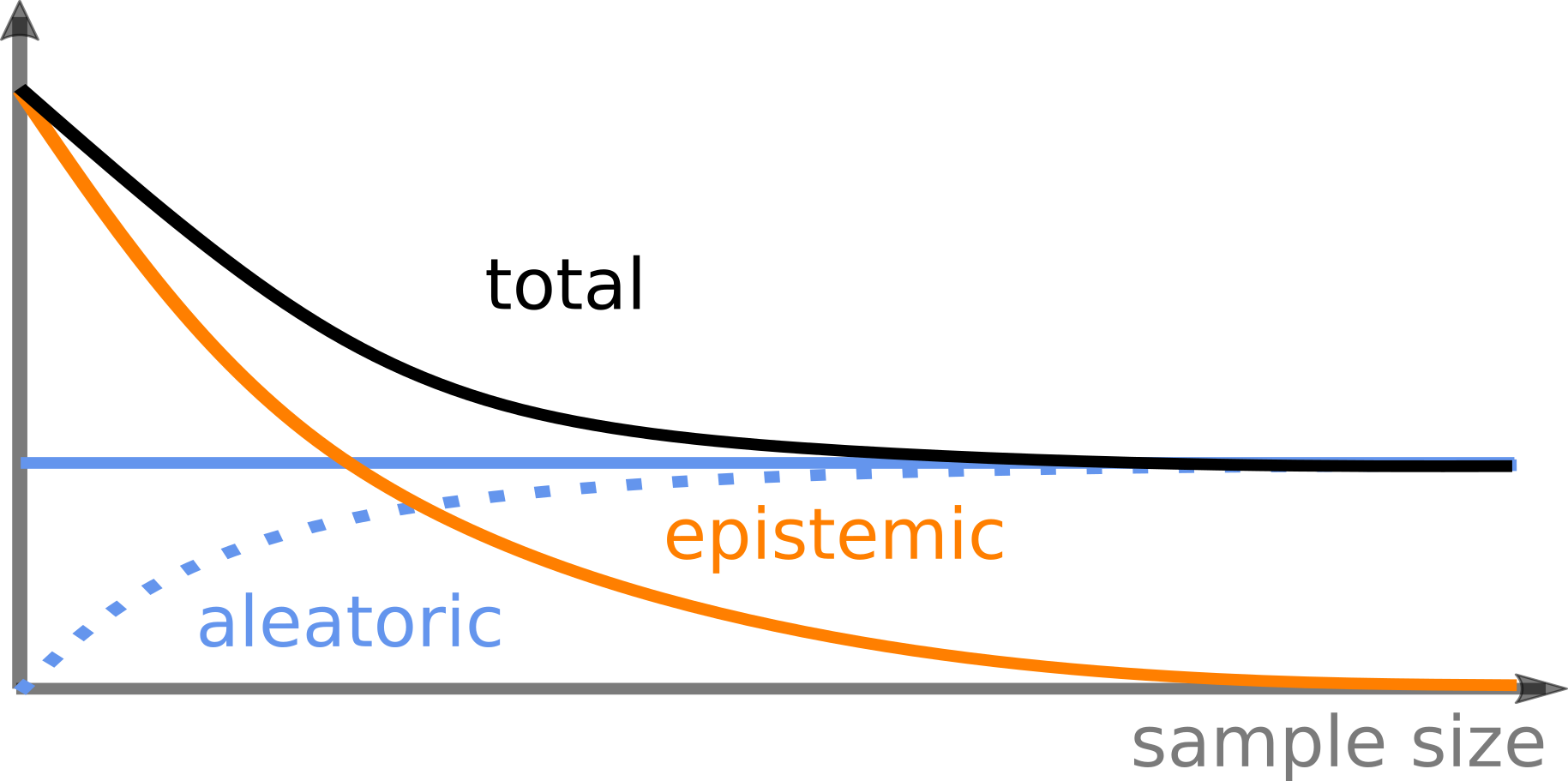}
        \caption{
        In the beginning, the learner is very uncertain about the true class probability for the (green) query point $\vec{x}$ (top-left), but this (epistemic) uncertainty disappears with increasing sample size (top-right). The bottom plot shows the expected qualitative behavior of uncertainty components. 
        The dotted line marks the difference of TU and EU (which equals AU when stipulating additivity).}
        \label{fig:decomp}
    \end{center}
\end{figure}

The dependence between the measures can be seen most clearly at the beginning of a learning process, during which the number of observed samples increases when the learner has seen very little or even no data. 
For example, take the case of binary classification, where the learner is still fully ignorant about the probability $\theta = \prob(y_1 \given \vec{x})$ of the positive class in a certain point $\vec{x}$: it might be a clear positive case ($\theta = 1$), a clear negative case ($\theta = 0$), but also everything in between. 
In this situation, an ideal measure of TU should arguably assume its maximum value. 
Likewise, an ideal measure of EU should be maximal because the learner is as uninformed about $\theta$ as it can be (note this is \emph{not} the case for mutual information when starting with a uniform prior). 
Then, however, additivity implies that AU must be zero, at least if the measures share a common scale with the same maximum value (which holds for entropy and mutual information). 
In a sense, the aleatoric part is already comprised in the epistemic part, suggesting that the difference between TU and EU can, at best, be interpreted as a \emph{lower bound} to the true AU. 
This is depicted graphically in Fig.~\ref{fig:decomp}: With a growing sample size, we expect EU to decrease and eventually vanish because the learner will gain full knowledge about the data-generating process, while both TU and TU minus EU should converge to the true AU (the former from above and the latter from below). 

    The above can be summarized as follows:
\begin{proposition}
    If EU and TU attain their respective maxima at the beginning of learning, and they are constructed to be on the same scale, then TU cannot decompose additively into EU and AU if AU is positive.
\end{proposition}

What these considerations suggest is that, in the finite-sample (machine learning) regime, additivity might inevitably be violated by \enquote{ideal} measures of TU, EU, and (ground-truth) AU (again, note that entropy and mutual information with a uniform prior are \emph{not} ideal in this sense, which is what we criticized them for). 
Instead, additivity may only hold for the decomposition of TU into EU and a \emph{lower bound} to AU, or might be relaxed to sub-additivity along the lines of \citet{dubois_1996_RepresentingPartialIgnorance}.
Alternatively, additivity must be preserved by giving up other assumptions, e.g., that all measures share the same scale. 
Recently, for example, \citet{hullermeier_2022_QuantificationCredalUncertainty} proposed a decomposition for the uncertainty of sets of probability distributions (credal sets), in which EU can become twice as high as AU. 

\section{Experiments}
\label{sec:experiments}

\subsection{Experimental Setup}

We conduct a number of experiments that provide practical evidence about the incoherent behavior of entropy-based uncertainty measures.
For further details on training configurations, see supplementary material \ref{app:expdetails}.
The experimental code is available in full via a public repository\footnote{\href{https://github.com/lisa-wm/entropybaseduq}{\texttt{https://github.com/lisa-wm/entropybaseduq}}}.

\paragraph{Datasets}
We consider image classification tasks for two real-world datasets,
\texttt{CIFAR10} \citep{Krizhevsky2009learning} and \texttt{MNIST} \citep{lecun_1998_GradientBasedLearningApplied}, containing ten classes of color and gray-scale images, respectively.
For a setting over which we can exert full control, we further synthesize black-and-white images of rectangles, where one class is characterized by vertical extension (i.e., height $>$ width) and the other \textit{vice versa}, and simulate a bivariate classification problem with tabular features and four classes.

\paragraph{Probabilistic Learners}

For the computer vision tasks, we employ deep ensembles of $M$ neural networks with varying random initialization \citep{lakshminarayanan_2017_SimpleScalablePredictive}, as well as a Laplace approximation \citep{ daxberger_2021_LaplaceReduxEffortless} that fits a Gaussian approximate posterior to the location of the maximum-a-posteriori estimator and draws $M$ samples from this distribution. 
Furthermore, we train a random forest \citep{breiman_2001_RandomForests} and an ensemble of single-hidden-layer feedforward neural networks (MLPs) for the tabular classification problem.
Ensemble predictions are computed by averaging over the outputs of individual members.
We set $M = 10$.
While this may seem small, note that this is in accordance with common practice (e.g., \citet{beluch_2018_PowerEnsemblesActive, lee_2021_SUNRISESimpleUnified, abe_2022_DeepEnsemblesWork, kristiadi_2022_BeingBitFrequentist, turkoglu_2022_FiLMEnsembleProbabilisticDeep}) because larger ensembles are rarely affordable in the deep learning with architectures typically numbering in the millions.
Our ablations suggest that the observed qualitative behavior is fairly robust with respect to ensemble size (see supplementary material \ref{app:nmembers}).  

\paragraph{Evaluation}
The figures in the following paragraphs visualize estimates for AU, EU, and TU (normalized to values in $[0, 1]$) alongside predictive performance in terms of accuracy (ACC) and expected calibration error (ECE; \citet{naeini_2015_ObtainingWellCalibrated}).
The uncertainty estimates are depicted as lines (averaged over all test samples, left $y$-axis), while the bars represent performance (right $y$-axis).
All results are aggregrated over three independent runs, with error bars on the uncertainty curves indicating one standard deviation. 

\subsection{Results and Discussion}
\subsubsection{Increasing Sample Size}

We first study how the components of total Shannon entropy evolve with sample size, where we admit increasingly larger training datasets from 1\% to 100\% of available samples (randomly selected; the classes are balanced).

\paragraph{Expected Behavior}
TU and EU start from their highest level and decrease gradually, while
AU approaches its true constant value from below (due to the additivity assumption).
The reported uncertainty is higher when accuracy is low.

\begin{figure}[H]
  \centering
      \includegraphics[width=0.4\textwidth]{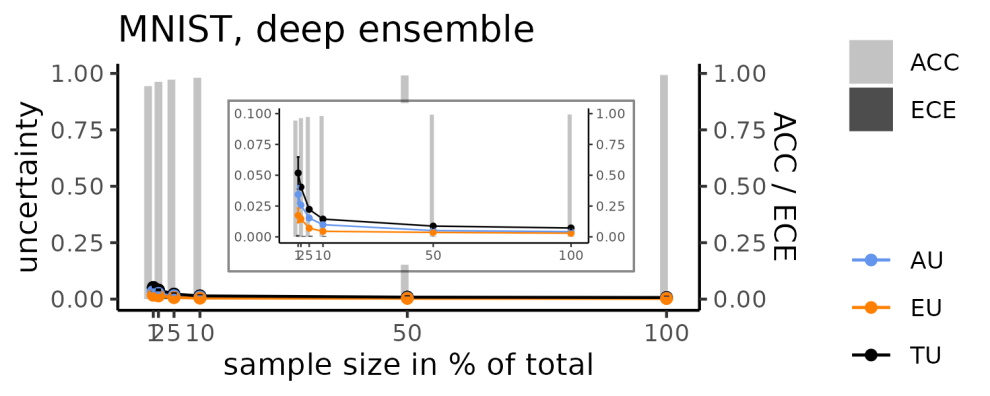}
      \includegraphics[width=0.4\textwidth]{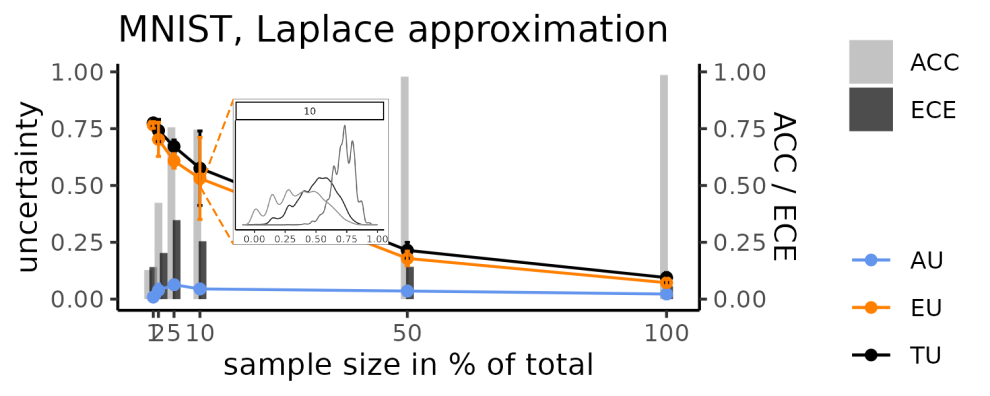}
  \caption{Entropy-based uncertainty for increasing sample size (\texttt{MNIST}). Overlaying plot (top): re-scaled left $y$-axis. Overlaying plot (bottom): empirical densities of observed instance-wise EU values (each curve represents one run of the experiment) with 10\% sample size.}
  \label{fig:uvssamples_mnist}
\end{figure}

\begin{figure}[H]
  \centering
      \includegraphics[width=0.4\textwidth]{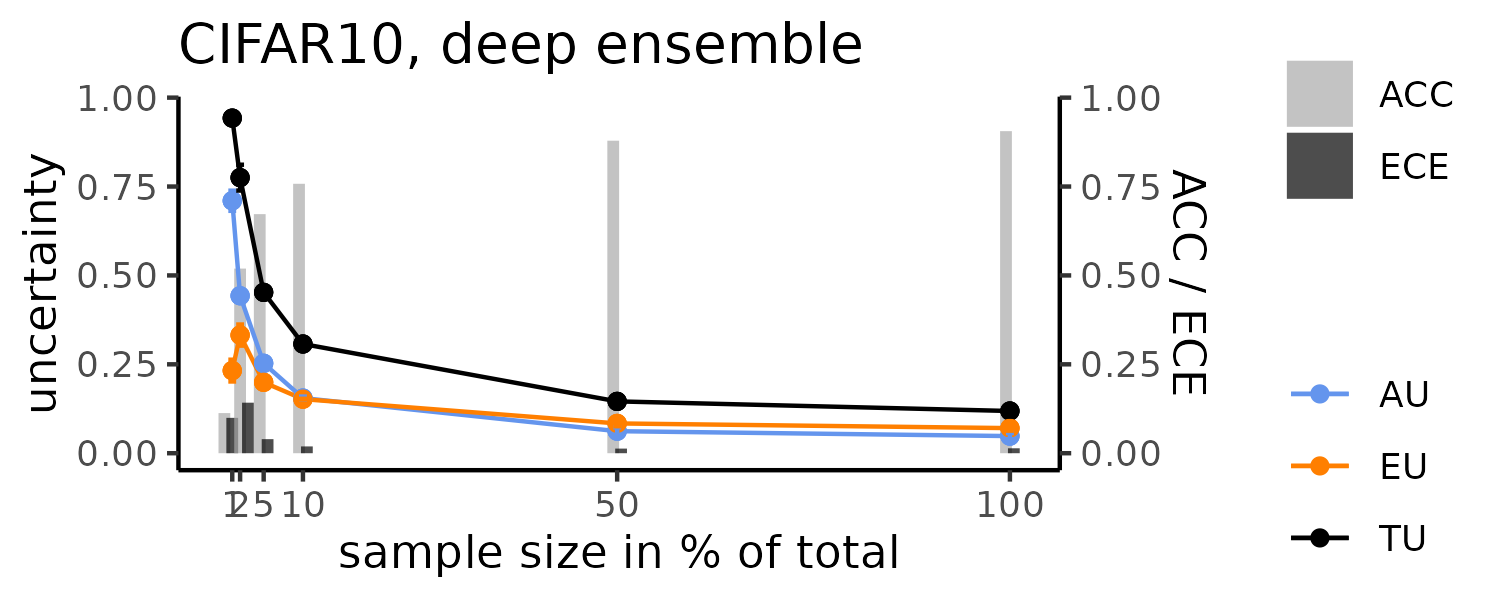}
      \includegraphics[width=0.4\textwidth]{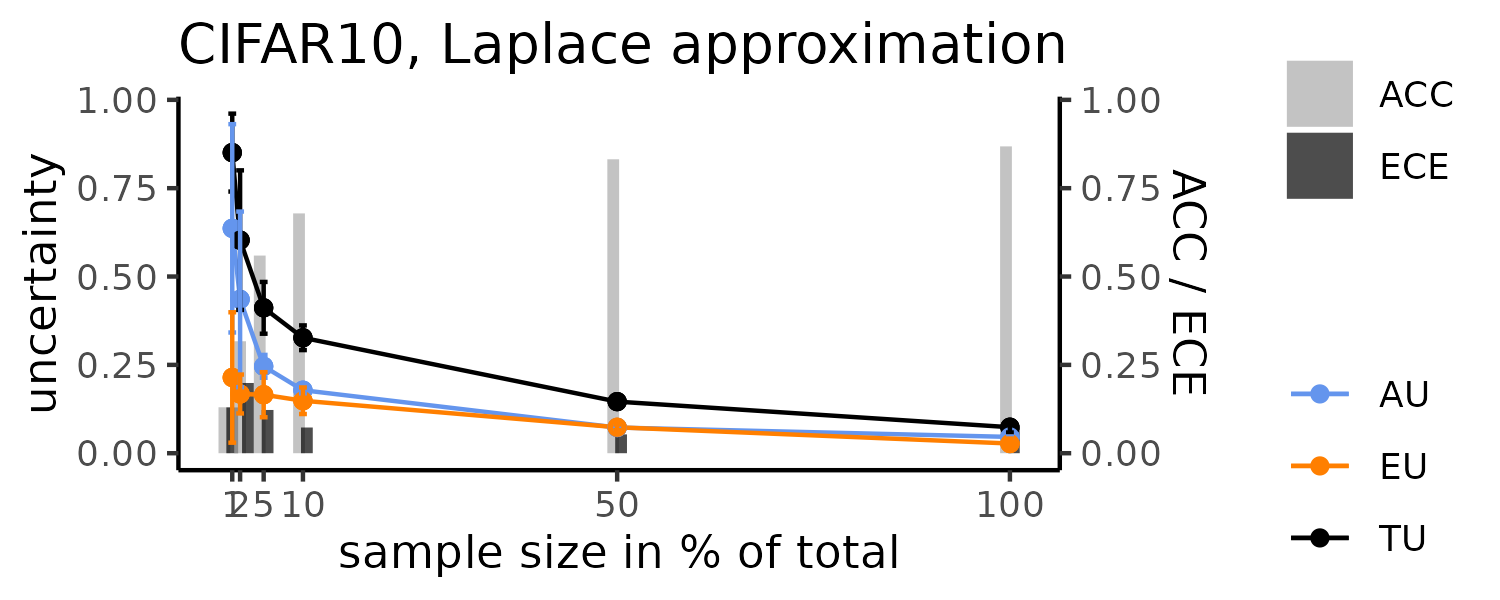}
  \caption{Entropy-based uncertainty for increasing sample size (\texttt{CIFAR10}).}
  \label{fig:uvssamples_cifar}
\end{figure}

\paragraph{Observed Behavior}
We observe the expected downward slope of TU and EU as a general pattern, but the evidence is otherwise not quite consistent.
Most strikingly, the AU estimates also assume high initial values and decrease for larger sample size in three out of four cases.
For \texttt{CIFAR10}, this behavior is clearly visible (Fig.~\ref{fig:uvssamples_cifar}), underlining our concern that AU estimates are unreliable in the light of limited knowledge.
In the case of \texttt{MNIST} and the deep ensemble, uncertainty is ultra-low from the beginning, but a close-up in the small overlaying plot of Fig.~\ref{fig:uvssamples_mnist} (top) reveals the same trend.
Reported EU for small sample sizes seems quite low overall, despite limited information and consequently poor performance.
A notable exception is the Laplace approximation for \texttt{MNIST} (Fig.~\ref{fig:uvssamples_mnist}, bottom), which resembles most closely what we would have expected, though it appears more susceptible to random changes (see close-up).

\subsubsection{Increasing Data Noise}

\begin{figure}[H]
    \centering
    \includegraphics[width=0.4\textwidth]{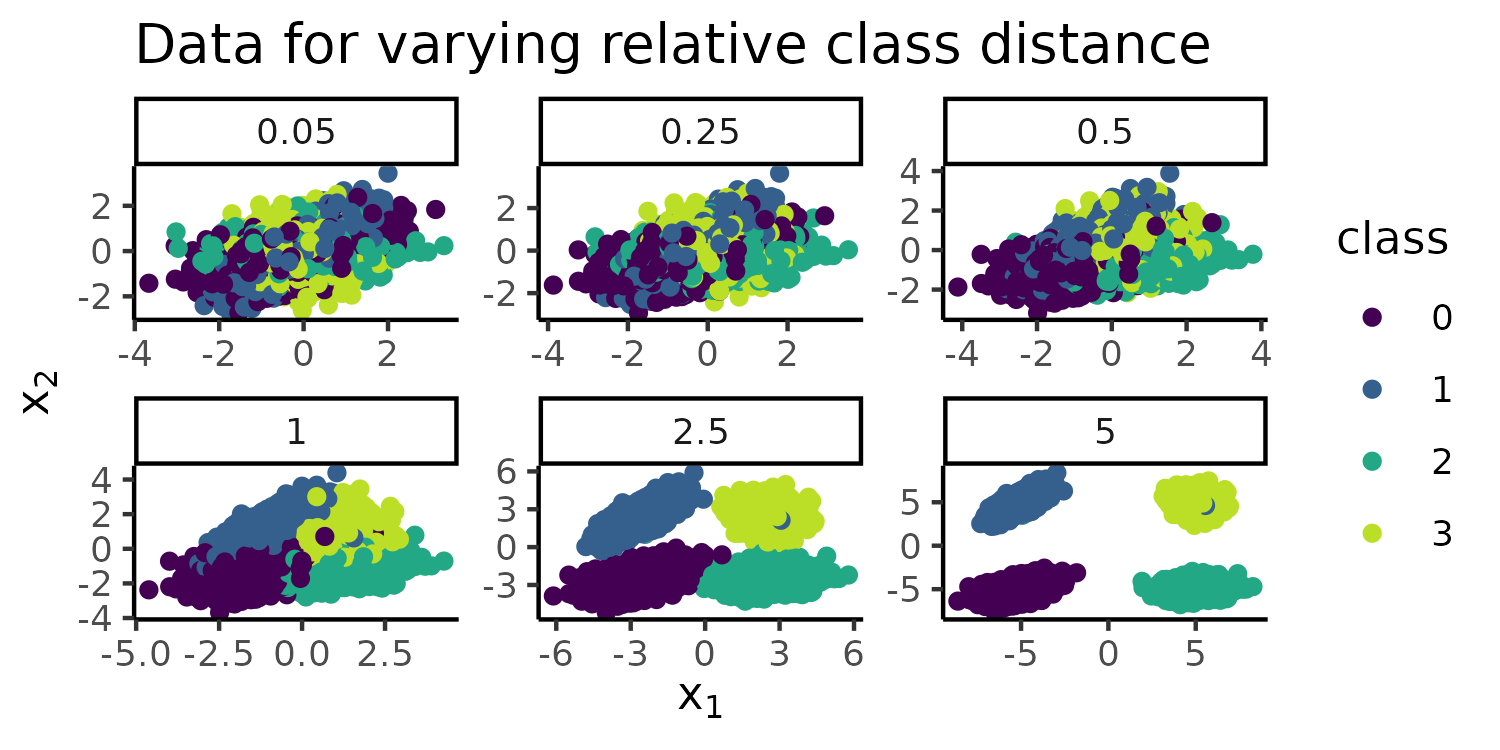}
    \caption{Tabular data with two features and four classes for increasing class overlap.}
    \label{fig:uvsdistance_data}
\end{figure}

Next, we modify the level of noise in the data, for which we use different proxies.
First, we vary image resolution in the computer vision tasks by downscaling (original sizes: 28$\times$28 / 32$\times$ 32 for \texttt{MNIST} / \texttt{CIFAR10}); second, we shrink the relative class distance for the tabular data and thus enforce stronger overlap (Fig.~\ref{fig:uvsdistance_data}); and lastly, we randomly change class labels for a varying share of observations in the same dataset (see supplementary material \ref{app:labelnoise}).

\paragraph{Expected Behavior}
AU picks up with increasing noise level.
Since learner capacity remains fixed, it is reasonable to assume that EU also rises to some extent when the decision boundaries become more complex with mounting degree of dataset contamination.

\begin{figure}[H]
  \centering
      \includegraphics[width=0.4\textwidth]{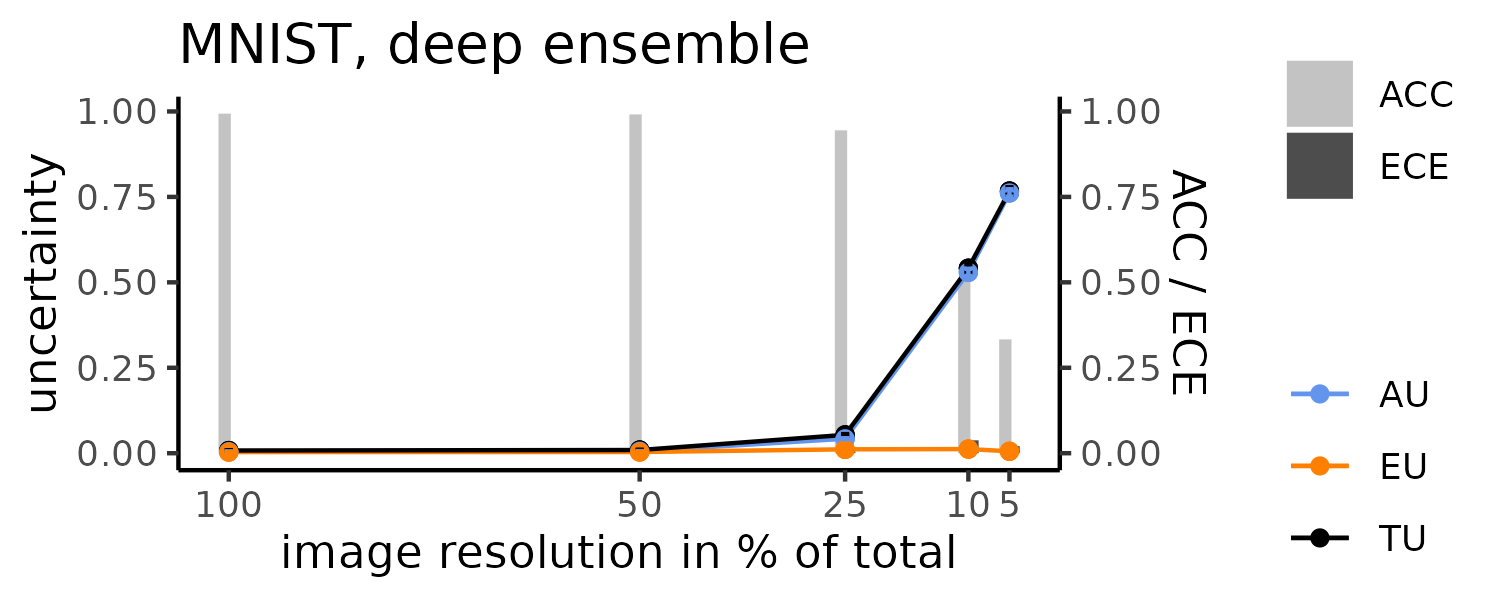}
      \includegraphics[width=0.4\textwidth]{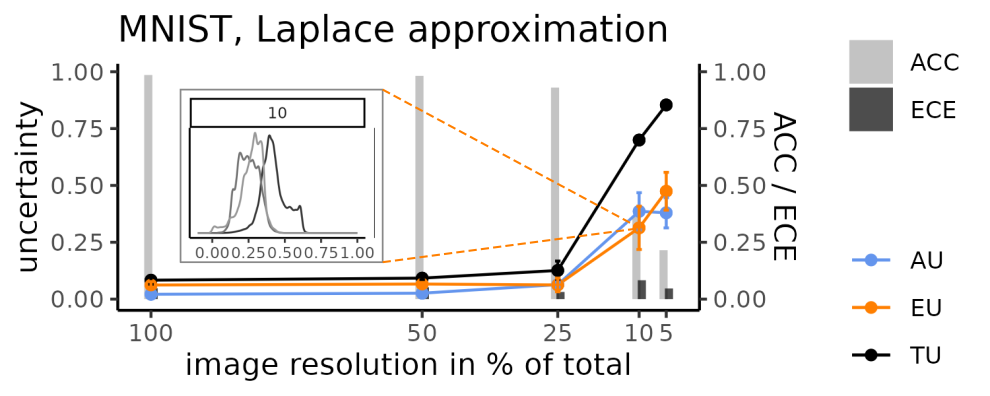}
  \caption{Entropy-based uncertainty for decreasing image resolution (\texttt{MNIST}).
  Overlaying plot: empirical densities of observed instance-wise EU values (each curve represents one run of the experiment) with 10\% resolution.}
  \label{fig:uvsresolution_mnist}
\end{figure}

\begin{figure}[H]
  \centering
      \includegraphics[width=0.4\textwidth]{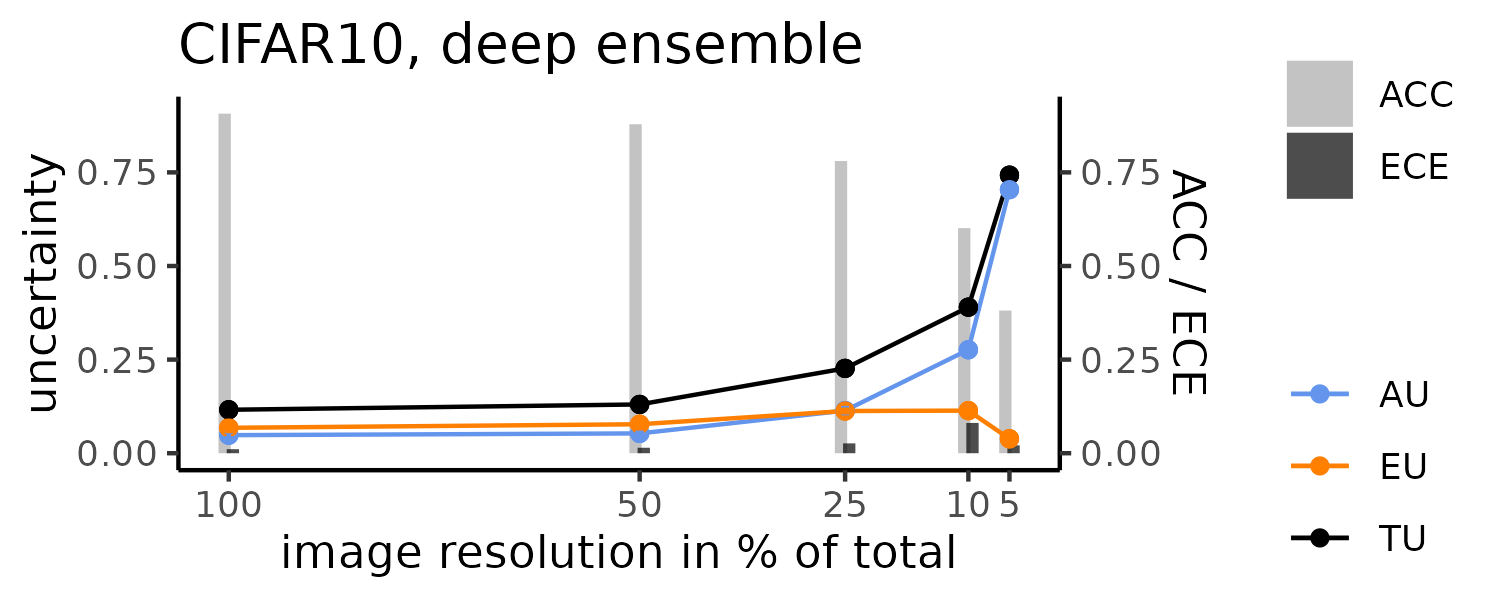}
      \includegraphics[width=0.4\textwidth]{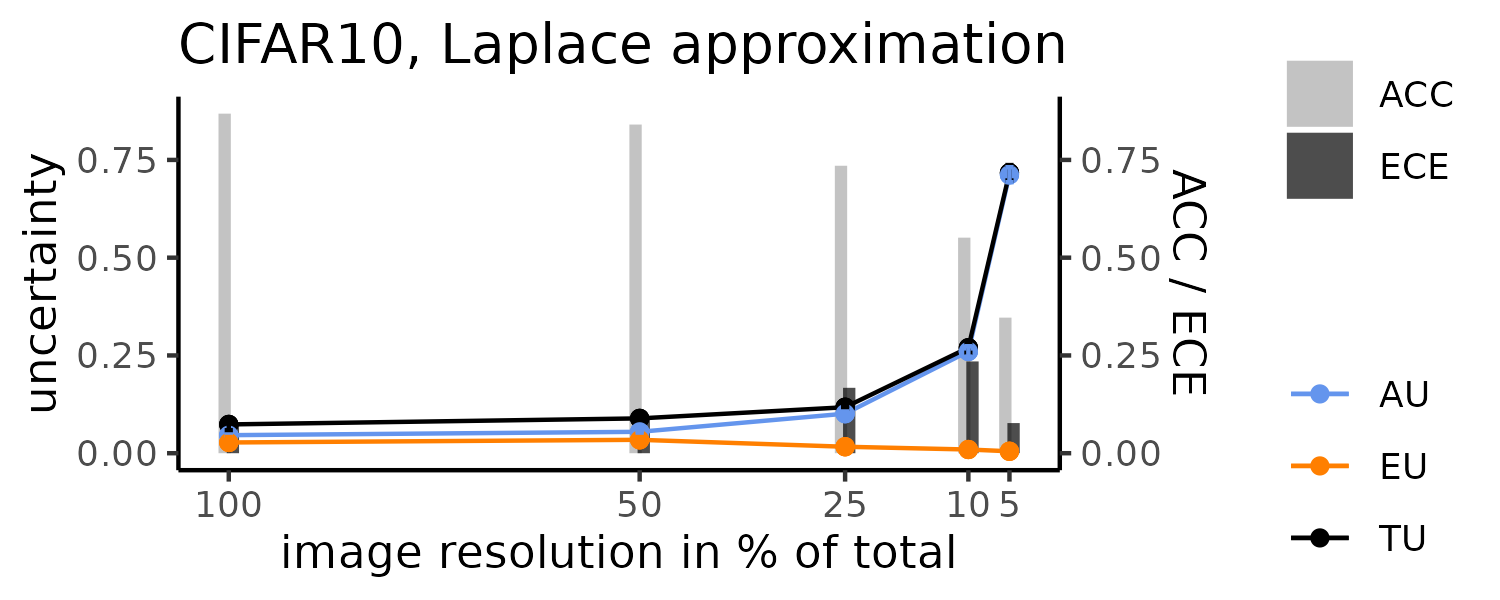}
  \caption{Entropy-based uncertainty for decreasing image resolution (\texttt{CIFAR10}).}
  \label{fig:uvsresolution_cifar}
\end{figure}

\begin{figure}[H]
    \centering
    \includegraphics[width=0.4\textwidth]{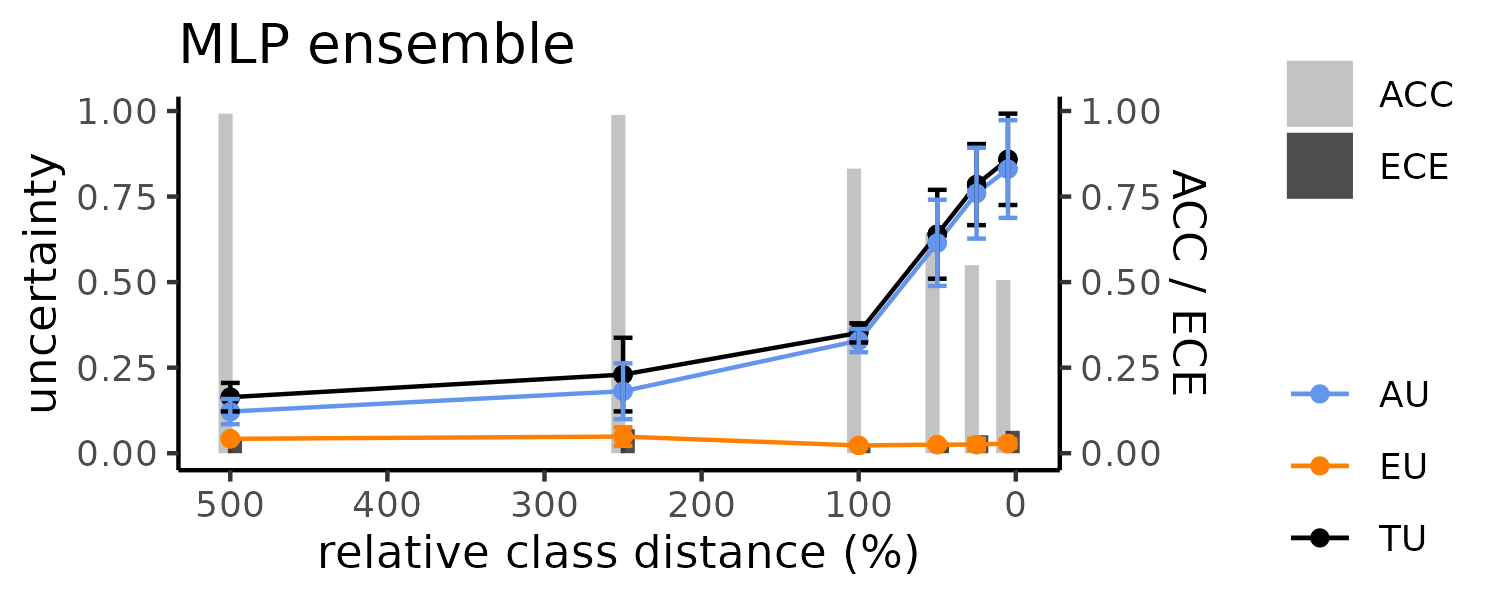}
    \caption{Entropy-based uncertainty for for increasing class overlap (tabular data).}
    \label{fig:uvsdistance_mlp}
\end{figure}

\paragraph{Observed Behavior}
More noise indeed prompts a rise in AU in all cases (Fig.~\ref{fig:uvsresolution_mnist}-\ref{fig:uvsdistance_mlp}).
However, EU remains basically constant and very low, even when the data are so noisy that performance plummets.
Note that, for instance, 10\% resolution corresponds to a compression into 3$\times$3 pixels for both \texttt{MNIST} and \texttt{CIFAR10}, and how the decision boundaries become quite blurred in the tabular dataset when classes overlap more strongly (Fig.~\ref{fig:uvsdistance_data}).  
Again, the Laplace approximation for \texttt{MNIST} is something of an exception: here, EU increases for the lowest resolution values, but leaves AU remarkably small and even falling from 10\% to 5\% resolution.
We find thus that the uncertainty measures behave, if not entirely implausible, again somewhat inconsistently.

\subsubsection{Test Distribution Shift}

\begin{minipage}[c]{0.1\textwidth}
    \begin{figure}[H]
             \includegraphics[width=0.75\textwidth]{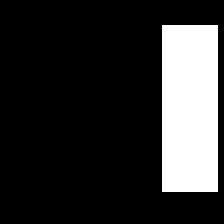}
             \includegraphics[width=0.75\textwidth]{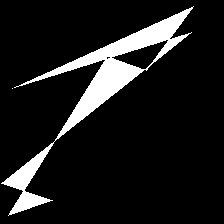}
    \end{figure}
    \vfill
\end{minipage}
\begin{minipage}[c]{0.38\textwidth}

Following a slightly different idea than in the previous experiment, we now train the learner on a well-separated dataset of synthetic rectangles and seek predictions for instances which interpolate between the classes.
This is achieved by decreasing the rectangles' side-length ratio, effectively pushing them towards a square.
\end{minipage}
 
We further present the learner with samples that are from an entirely different class, namely random non-convex polygons with 3--5 vertices, and thus out-of-distribution (OOD) relative to the training data.
Both settings create a form of distribution shift between training and test data that is associated with the presence of predictive uncertainty.

\paragraph{Expected Behavior}
Class interpolation mainly affects AU through instances located close to the learned decision boundaries.
In OOD detection, demanding predictions for samples from a previously unseen class should spike an increase in EU.

\begin{figure}[H]
  \centering
      \includegraphics[width=0.45\textwidth]{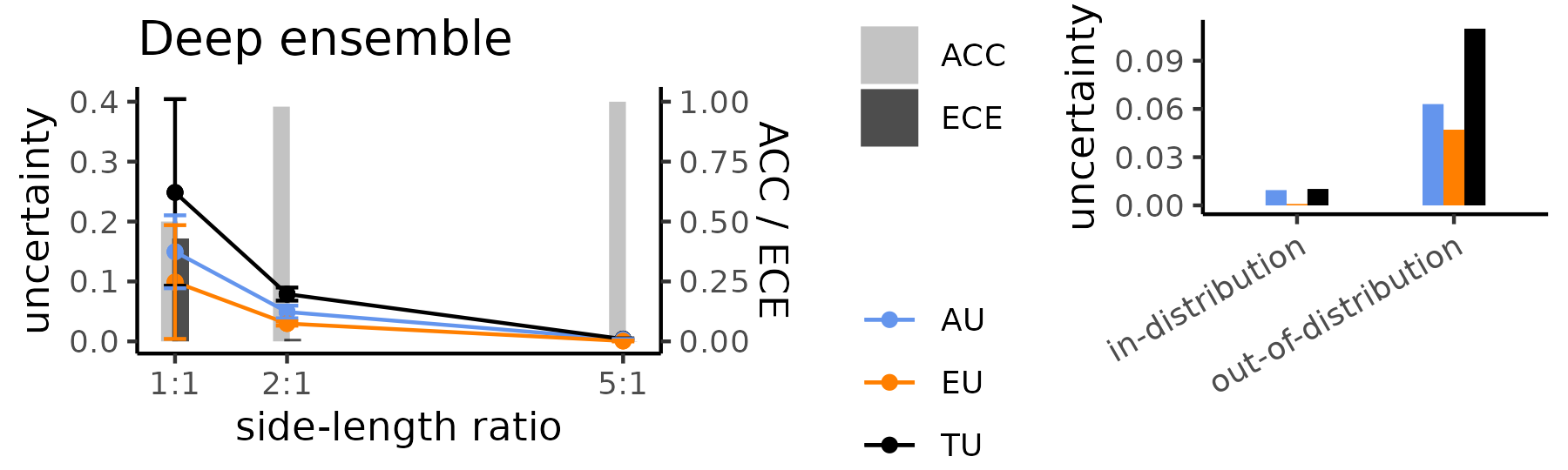}
      \includegraphics[width=0.45\textwidth]{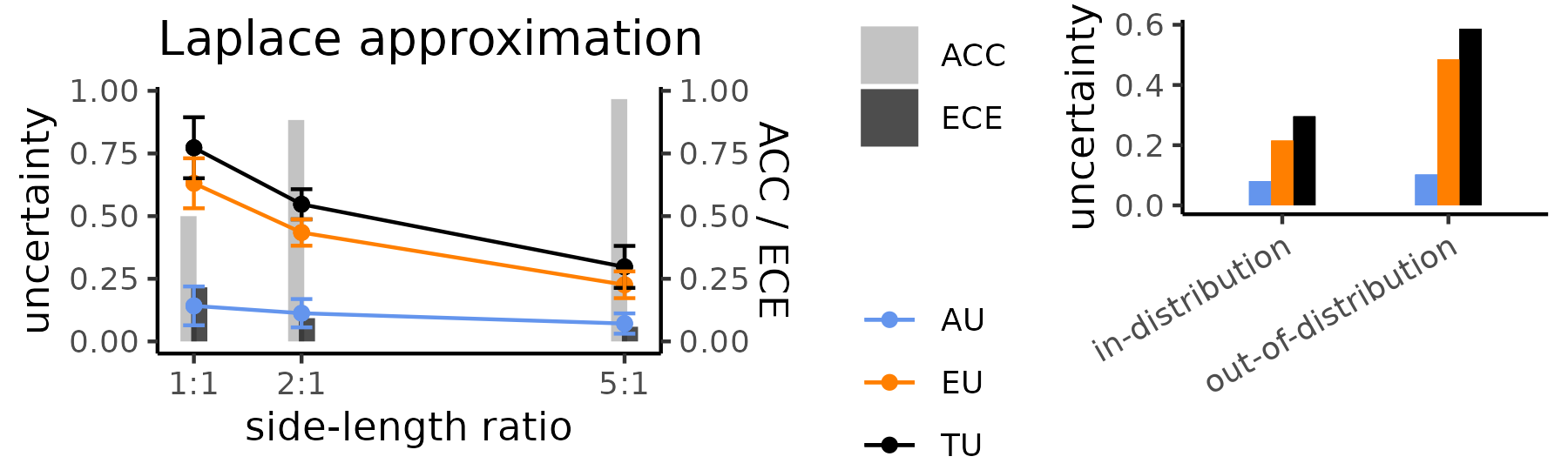}
  \caption{Entropy-based uncertainty for synthetic data with increasing class interpolation (left) and OOD detection (random polygons; right).}
  \label{fig:uvssynth}
\end{figure}

\paragraph{Observed Behavior}
The deep ensemble and Laplace approximation act quite differently in the presence of class interpolation (Fig.~\ref{fig:uvssynth}). 
While the former behaves roughly in line with our expectation, the Laplace learner attributes most of the uncertainty to the epistemic component when samples become increasingly quadratic, leaving AU almost unchanged.
In OOD detection, both classifiers react appropriately in principle by reporting higher uncertainty for the polygon images.
However, they allocate the majority of the uncertainty premium differently: Laplace almost exclusively raises the epistemic component, the deep ensemble signals both higher AU and EU.
In both ablations, a vast gap in overall levels of uncertainty is evident (e.g., for the OOD case, Laplace reports roughly five times higher TU).

\subsubsection{Discussion}

Our experiments confirm empirically that the entropy-based uncertainty measures exhibit some -- sometimes subtle -- inconsistencies in different learning situations.
While TU as a whole behaves fairly sensibly, the attribution to its aleatoric and epistemic components often appears implausible.
We further identify very low levels of EU as a global pattern.
That said, we also find that the design of the probabilistic learner used can affect the results considerably, even when predictive performance remains similar.
In particular, deep ensembles draw from the opinions of various base learners when they estimate predictive uncertainty, as opposed to Laplace approximation, where a single base network determines uncertainty quantification.
Consequently, the behavior of explicit ensembles also depends on the complexity of their base learners in that AU estimates decrease considerably when each ensemble member is granted more capacity (see supplementary material \ref{app:complexity}).
This relates to another insight relevant for this kind of empirical studies: reasoning about uncertainty estimates must always occur in conjunction with predictive accuracy and calibration, since we can hardly expect sensible uncertainty behavior when the learner fails on the basic underlying task.
Numerous practical aspects thus interact to elicit or mitigate shortcomings of the discussed measures and warrant caution in interpreting any predictive uncertainty estimate one may obtain in practice.
Worryingly, this becomes all the more important with limited data, a situation often encountered in, e.g., medical applications.

\section{Concluding Remarks}

Despite the common use of Shannon entropy, conditional entropy, and mutual information for the quantification of predictive uncertainty, we observed and empirically confirmed that these measures behave neither coherently nor always as expected. 
First, each of the measures itself can be criticized for (the lack of) certain properties. 
For example, we argued that mutual information is rather a measure of divergence than a measure of ignorance. 
Second, the discrepancy between (i) the ground-truth scenario concerning the true data-generating process with an objective ground-truth AU and zero EU, and (ii) the finite-sample setting in which AU is \emph{derived} from (and hence depends on) subjective EU, leads to incoherencies and calls additivity of the decomposition into question.
With this in mind, we encourage a critical assessment of the existing framework and cautious use of these measures.  

When remaining committed to the basic (Bayesian) representational framework with second-order distributions for the epistemic part of uncertainty, an obvious alternative is to look for measures with better properties, and indeed, some proposals can be found in the recent literature (e.g., \citet{DBLP:journals/corr/abs-2012-12687}). 
Another option is to change not only the measures for quantification, but also the underlying representation, for example by moving beyond classical probability. 
In fact, the suitability of probability distributions to represent ignorance in the sense of a lack of knowledge has been questioned by various scholars \citep{dubois_1996_RepresentingPartialIgnorance}, who advocate the use of generalized, more expressive uncertainty formalisms. 
In particular, it has been argued that the uniform distribution is not a suitable representation of complete ignorance, as it fails to distinguish the epistemic state of zero knowledge and the state of perfect information, where the learner knows that all outcomes are equiprobable. 
Indeed, the averaging over conditional entropies, which is part of the problems with the aleatoric component, is arguably meaningful in the latter case but much less so in the former. 
Seen from this perspective, one may indeed wonder whether classical probability is the right paradigm for modeling the epistemic part. On the other side, measuring uncertainty and disaggregating total predictive uncertainty into its aleatoric and epistemic components do not necessarily become simpler for generalized formalisms \citep{hullermeier_2022_QuantificationCredalUncertainty}. 
Clearly, we are not yet at the end of the path toward a truly meaningful uncertainty representation and quantification.

\begin{acknowledgements}
This work was supported by the DAAD program Konrad Zuse Schools of Excellence in Artificial Intelligence, sponsored by the Federal Ministry of Education and Research (LW, YS).
\end{acknowledgements}

\nocite{tan_2019_EfficientNetRethinkingModel, scikit-learn, paszke_2019_PyTorchImperativeStyle, daxberger_2021_LaplaceReduxEffortless, lightningai_2023_PyTorchLightningTrain}


\begin{thebibliography}{53}
\providecommand{\natexlab}[1]{#1}
\providecommand{\url}[1]{\texttt{#1}}
\expandafter\ifx\csname urlstyle\endcsname\relax
  \providecommand{\doi}[1]{doi: #1}\else
  \providecommand{\doi}{doi: \begingroup \urlstyle{rm}\Url}\fi

\bibitem[Abe et~al.(2022)Abe, Buchanan, Pleiss, Zemel, and
  Cunningham]{abe_2022_DeepEnsemblesWork}
T.~Abe, E.~K. Buchanan, G.~Pleiss, R.~Zemel, and J.~P. Cunningham.
\newblock Deep {{Ensembles Work}}, {{But Are They Necessary}}?, 2022.

\bibitem[Andrieu et~al.(2003)Andrieu, De~Freitas, Doucet, and
  Jordan]{andrieu_2003_IntroductionMCMCMachine}
C.~Andrieu, N.~De~Freitas, A.~Doucet, and M.~I. Jordan.
\newblock An {{Introduction}} to {{MCMC}} for {{Machine Learning}}.
\newblock \emph{Machine Learning}, 50:\penalty0 5--43, 2003.

\bibitem[Ash(1965)]{ash_1965_InformationTheory}
R.~B. Ash.
\newblock \emph{Information {{Theory}}}.
\newblock {Dover Publications}, 1965.

\bibitem[Ashukha et~al.(2020)Ashukha, Lyzhov, Molchanov, and
  Vetrov]{ashukha_2020_PitfallsInDomainUncertaintya}
A.~Ashukha, A.~Lyzhov, D.~Molchanov, and D.~Vetrov.
\newblock Pitfalls of {{In-Domain Uncertainty Estimation}} and {{Ensembling}}
  in {{Deep Learning}}.
\newblock In \emph{{{ICLR}}}, 2020.

\bibitem[Beluch et~al.(2018)Beluch, Genewein, Nurnberger, and
  Kohler]{beluch_2018_PowerEnsemblesActive}
W.~H. Beluch, T.~Genewein, A.~Nurnberger, and J.~M. Kohler.
\newblock The {{Power}} of {{Ensembles}} for {{Active Learning}} in {{Image
  Classification}}.
\newblock In \emph{{{CVF Conference}} on {{Computer Vision}} and {{Pattern
  Recognition}}}. {IEEE}, 2018.

\bibitem[Breiman(2001)]{breiman_2001_RandomForests}
L.~Breiman.
\newblock Random {{Forests}}.
\newblock \emph{Machine Learning}, 45\penalty0 (1):\penalty0 5--32, 2001.

\bibitem[Bronevich and Klir(2008)]{bronevich2008axioms}
A.~Bronevich and G.~J. Klir.
\newblock Axioms for uncertainty measures on belief functions and credal sets.
\newblock In \emph{NAFIPS}, pages 1--6. IEEE, 2008.

\bibitem[Budd et~al.(2021)Budd, Robinson, and
  Kainz]{budd_2021_SurveyActiveLearning}
S.~Budd, E.~C. Robinson, and B.~Kainz.
\newblock A {{Survey}} on {{Active Learning}} and {{Human-in-the-Loop Deep
  Learning}} for {{Medical Image Analysis}}.
\newblock \emph{Medical Image Analysis}, 71, 2021.

\bibitem[Charpentier et~al.(2020)Charpentier, Zügner, and
  Günnemann]{charpentier_2020_PosteriorNetworkUncertainty}
B.~Charpentier, D.~Zügner, and S.~Günnemann.
\newblock Posterior {{Network}}: {{Uncertainty Estimation}} without {{OOD
  Samples}} via {{Density-Based Pseudo-Counts}}.
\newblock In \emph{{{NeurIPS}}}, pages 1103--1130, 2020.

\bibitem[Charpentier et~al.(2022)Charpentier, Senanayake, Kochenderfer, and
  Günnemann]{charpentier_2022_DisentanglingEpistemicAleatoric}
B.~Charpentier, R.~Senanayake, M.~Kochenderfer, and S.~Günnemann.
\newblock Disentangling {{Epistemic}} and {{Aleatoric Uncertainty}} in
  {{Reinforcement Learning}}, 2022.

\bibitem[Cover and Thomas(2006)]{cover_2006_ElementsInformationTheory}
T.~M. Cover and J.~A. Thomas.
\newblock \emph{Elements of {{Information}} {{Theory}}}.
\newblock {Wiley}, 2 edition, 2006.

\bibitem[Daxberger et~al.(2021)Daxberger, Kristiadi, Immer, Eschenhagen, Bauer,
  and Hennig]{daxberger_2021_LaplaceReduxEffortless}
E.~Daxberger, A.~Kristiadi, A.~Immer, R.~Eschenhagen, M.~Bauer, and P.~Hennig.
\newblock Laplace {{Redux}} – {{Effortless Bayesian Deep Learning}}, 2021.

\bibitem[Depeweg(2019)]{depeweg_2019_ModelingEpistemicAleatoric}
S.~Depeweg.
\newblock \emph{Modeling {{Epistemic}} and {{Aleatoric Uncertainty}} with
  {{Bayesian Neural Networks}} and {{Latent Variable}}}.
\newblock PhD thesis, {TU München}, 2019.

\bibitem[Depeweg et~al.(2018)Depeweg, Hernández-Lobato, Doshi-Velez, and
  Udluft]{depeweg_2018_DecompositionUncertaintyBayesian}
S.~Depeweg, J.~M. Hernández-Lobato, F.~Doshi-Velez, and S.~Udluft.
\newblock Decomposition of {{Uncertainty}} in {{Bayesian Deep Learning}} for
  {{Efficient}} and {{Risk-sensitive Learning}}, 2018.

\bibitem[Dubois et~al.(1996)Dubois, Prade, and
  Smets]{dubois_1996_RepresentingPartialIgnorance}
D.~Dubois, H.~M. Prade, and P.~Smets.
\newblock Representing partial ignorance.
\newblock \emph{IEEE Transactions on Systems, Man, and Cybernetics - Part A:
  Systems and Humans}, 26\penalty0 (3):\penalty0 361--377, 1996.

\bibitem[Gal(2016)]{gal_2016_UncertaintyDeepLearning}
Y.~Gal.
\newblock \emph{Uncertainty in {{Deep Learning}}}.
\newblock PhD thesis, {University of Cambridge}, 2016.

\bibitem[Gal et~al.(2017)Gal, Islam, and
  Ghahramani]{gal_2017_DeepBayesianActivea}
Y.~Gal, R.~Islam, and Z.~Ghahramani.
\newblock Deep {{Bayesian Active Learning}} with {{Image Data}}, 2017.

\bibitem[Gelman et~al.(2021)Gelman, Carlin, Stern, Dunson, Vehtari, and
  Rubin]{gelman_2021_BayesianDataAnalysis}
A.~Gelman, J.~Carlin, H.~Stern, D.~Dunson, A.~Vehtari, and D.~Rubin.
\newblock \emph{Bayesian {{Data Analysis}}}.
\newblock {CRC Press}, 2021.

\bibitem[Houlsby et~al.(2011)Houlsby, Huszár, Ghahramani, and
  Lengyel]{houlsby_2011_BayesianActiveLearning}
N.~Houlsby, F.~Huszár, Z.~Ghahramani, and M.~Lengyel.
\newblock Bayesian {{Active Learning}} for {{Classification}} and {{Preference
  Learning}}, 2011.

\bibitem[Huseljic et~al.(2020)Huseljic, Sick, Herde, and
  Kottke]{huseljic_2020_SeparationAleatoricEpistemic}
D.~Huseljic, B.~Sick, M.~Herde, and D.~Kottke.
\newblock Separation of {{Aleatoric}} and {{Epistemic Uncertainty}} in
  {{Deterministic Deep Neural Networks}}.
\newblock In \emph{{{ICPR}}}, 2020.

\bibitem[Hüllermeier and
  Waegeman(2021)]{hullermeier_2021_AleatoricEpistemicUncertainty}
E.~Hüllermeier and W.~Waegeman.
\newblock Aleatoric and {{Epistemic Uncertainty}} in {{Machine Learning}}: {{An
  Introduction}} to {{Concepts}} and {{Methods}}.
\newblock \emph{Machine Learning}, 2021.

\bibitem[Hüllermeier et~al.(2022)Hüllermeier, Destercke, and
  Shaker]{hullermeier_2022_QuantificationCredalUncertainty}
E.~Hüllermeier, S.~Destercke, and M.~H. Shaker.
\newblock Quantification of {{Credal Uncertainty}} in {{Machine Learning}}: {{A
  Critical Analysis}} and {{Empirical Comparison}}.
\newblock In \emph{{{UAI}}}, 2022.

\bibitem[Kendall and Gal(2017)]{kendall_2017_WhatUncertaintiesWe}
A.~Kendall and Y.~Gal.
\newblock What {{Uncertainties Do We Need}} in {{Bayesian Deep Learning}} for
  {{Computer Vision}}?
\newblock In \emph{{{NeurIPS}}}, 2017.

\bibitem[Kirsch et~al.(2022)Kirsch, Kossen, and
  Gal]{kirsch_2022_MarginalJointCrossEntropies}
A.~Kirsch, J.~Kossen, and Y.~Gal.
\newblock Marginal and {{Joint Cross-Entropies}} \& {{Predictives}} for
  {{Online Bayesian Inference}}, {{Active Learning}}, and {{Active Sampling}},
  2022.

\bibitem[Kopetzki et~al.(2021)Kopetzki, Charpentier, Zügner, Giri, and
  Günnemann]{kopetzki_2021_EvaluatingRobustnessPredictive}
A.-K. Kopetzki, B.~Charpentier, D.~Zügner, S.~Giri, and S.~Günnemann.
\newblock Evaluating {{Robustness}} of {{Predictive Uncertainty Estimation}}:
  {{Are Dirichlet-based Models Reliable}}?
\newblock In \emph{{{ICML}}}, pages 5707--5718, 2021.

\bibitem[Kristiadi et~al.(2022)Kristiadi, Hein, and
  Hennig]{kristiadi_2022_BeingBitFrequentist}
A.~Kristiadi, M.~Hein, and P.~Hennig.
\newblock Being a {{Bit Frequentist Improves Bayesian Neural Networks}}.
\newblock In \emph{{{AISTATS}}}, 2022.

\bibitem[Krizhevsky(2009)]{Krizhevsky2009learning}
A.~Krizhevsky.
\newblock Learning multiple layers of features from tiny images.
\newblock Technical report, University of Toronto, 2009.

\bibitem[Lakshminarayanan et~al.(2017)Lakshminarayanan, Pritzel, and
  Blundell]{lakshminarayanan_2017_SimpleScalablePredictive}
B.~Lakshminarayanan, A.~Pritzel, and C.~Blundell.
\newblock Simple and {{Scalable Predictive Uncertainty Estimation}} using
  {{Deep Ensembles}}.
\newblock In \emph{{{NeurIPS}}}, 2017.

\bibitem[LeCun et~al.(1998)LeCun, Bottou, Bengio, and
  Haffner]{lecun_1998_GradientBasedLearningApplied}
Y.~LeCun, L.~Bottou, Y.~Bengio, and P.~Haffner.
\newblock Gradient-{{Based Learning Applied}} to {{Document Recognition}}.
\newblock In \emph{Proceedings of the {{IEEE}}}, 1998.

\bibitem[Lee et~al.(2021)Lee, Laskin, Srinivas, and
  Abbeel]{lee_2021_SUNRISESimpleUnified}
K.~Lee, M.~Laskin, A.~Srinivas, and P.~Abbeel.
\newblock {{SUNRISE}}: {{A Simple Unified Framework}} for {{Ensemble Learning}}
  in {{Deep Reinforcement Learning}}.
\newblock In \emph{{{ICML}}}, 2021.

\bibitem[{Lightning AI}(2023)]{lightningai_2023_PyTorchLightningTrain}
{Lightning AI}.
\newblock {{PyTorch Lightning}}: {{Train}} and {{Deploy PyTorch}} at {{Scale}},
  2023.

\bibitem[Malinin(2019)]{malinin_2019_UncertaintyEstimationDeep}
A.~Malinin.
\newblock \emph{Uncertainty {{Estimation}} in {{Deep Learning}} with
  {{Application}} to {{Spoken Language Assessment}}}.
\newblock PhD thesis, {University of Cambridge}, 2019.

\bibitem[Malinin and Gales(2018)]{malinin_2018_PredictiveUncertaintyEstimation}
A.~Malinin and M.~Gales.
\newblock Predictive {{Uncertainty Estimation}} via {{Prior Networks}}.
\newblock In \emph{{{NeurIPS}}}, 2018.

\bibitem[Michelmore et~al.(2018)Michelmore, Kwiatkowska, and
  Gal]{michelmore_2018_EvaluatingUncertaintyQuantification}
R.~Michelmore, M.~Kwiatkowska, and Y.~Gal.
\newblock Evaluating {{Uncertainty Quantification}} in {{End-to-End Autonomous
  Driving Control}}, 2018.

\bibitem[Mobiny et~al.(2021)Mobiny, Yuan, Moulik, Garg, Wu, and
  Van~Nguyen]{mobiny_2021_DropConnectEffectiveModeling}
A.~Mobiny, P.~Yuan, S.~K. Moulik, N.~Garg, C.~C. Wu, and H.~Van~Nguyen.
\newblock {{DropConnect}} is effective in modeling uncertainty of {{Bayesian}}
  deep networks.
\newblock \emph{Scientific Reports}, 11, 2021.

\bibitem[Naeini et~al.(2015)Naeini, Cooper, and
  Hauskrecht]{naeini_2015_ObtainingWellCalibrated}
M.~P. Naeini, G.~Cooper, and M.~Hauskrecht.
\newblock Obtaining {{Well Calibrated Probabilities Using Bayesian Binning}}.
\newblock In \emph{Proceedings of the {{AAAI Conference}} on {{Artificial
  Intelligence}}}, volume~29, 2015.

\bibitem[Ovadia et~al.(2019)Ovadia, Fertig, Ren, Nado, Sculley, Nowozin,
  Dillon, Lakshminarayanan, and Snoek]{ovadia_can_2019}
Y.~Ovadia, E.~Fertig, J.~Ren, Z.~Nado, D.~Sculley, S.~Nowozin, J.~V. Dillon,
  B.~Lakshminarayanan, and J.~Snoek.
\newblock Can {{You Trust Your Model}}'s {{Uncertainty}}? {{Evaluating
  Predictive Uncertainty Under Dataset Shift}}.
\newblock In \emph{Advances in {{Neural Information Processing Systems}} 32
  ({{NeurIPS}} 2019)}, 2019.

\bibitem[Pal et~al.(1992)Pal, Bezdek, and
  Hemasinha]{pal_1992_UncertaintyMeasuresEvidential}
N.~R. Pal, J.~C. Bezdek, and R.~Hemasinha.
\newblock Uncertainty measures for evidential reasoning {{I}}: {{A}} review.
\newblock \emph{International Journal of Approximate Reasoning}, 7\penalty0
  (3-4):\penalty0 165--183, 1992.

\bibitem[Pal et~al.(1993)Pal, Bezdek, and Hemasinha]{pal1993uncertainty}
N.~R. Pal, J.~C. Bezdek, and R.~Hemasinha.
\newblock Uncertainty measures for evidential reasoning ii: A new measure of
  total uncertainty.
\newblock \emph{International Journal of Approximate Reasoning}, 8\penalty0
  (1):\penalty0 1--16, 1993.

\bibitem[Paszke et~al.(2019)Paszke, Gross, Massa, Lerer, Bradbury, Chanan,
  Killeen, Lin, Gimelshein, Antiga, Desmaison, Kopf, Yang, DeVito, Raison,
  Tejani, Chilamkurthy, Steiner, Fang, Bai, and
  Chintala]{paszke_2019_PyTorchImperativeStyle}
A.~Paszke, S.~Gross, F.~Massa, A.~Lerer, J.~Bradbury, G.~Chanan, T.~Killeen,
  Z.~Lin, N.~Gimelshein, L.~Antiga, A.~Desmaison, A.~Kopf, E.~Yang, Z.~DeVito,
  M.~Raison, A.~Tejani, S.~Chilamkurthy, B.~Steiner, L.~Fang, J.~Bai, and
  S.~Chintala.
\newblock {{PyTorch}}: {{An Imperative Style}}, {{High-Performance Deep
  Learning Library}}.
\newblock In \emph{Proceedings of the 33rd {{Conference}} on {{Neural
  Information Processing Systems}} ({{NeurIPS}} 2019)}, 2019.

\bibitem[Pedregosa et~al.(2011)Pedregosa, Varoquaux, Gramfort, Michel, Thirion,
  Grisel, Blondel, Prettenhofer, Weiss, Dubourg, Vanderplas, Passos,
  Cournapeau, Brucher, Perrot, and Duchesnay]{scikit-learn}
F.~Pedregosa, G.~Varoquaux, A.~Gramfort, V.~Michel, B.~Thirion, O.~Grisel,
  M.~Blondel, P.~Prettenhofer, R.~Weiss, V.~Dubourg, J.~Vanderplas, A.~Passos,
  D.~Cournapeau, M.~Brucher, M.~Perrot, and E.~Duchesnay.
\newblock Scikit-learn: Machine learning in {P}ython.
\newblock \emph{Journal of Machine Learning Research}, 12:\penalty0 2825--2830,
  2011.

\bibitem[Psaros et~al.(2022)Psaros, Meng, Zou, Guo, and
  Karniadakis]{psaros_2022_UncertaintyQuantificationScientificb}
A.~F. Psaros, X.~Meng, Z.~Zou, L.~Guo, and G.~E. Karniadakis.
\newblock Uncertainty {{Quantification}} in {{Scientific Machine Learning}}:
  {{Methods}}, {{Metrics}}, and {{Comparisons}}, 2022.

\bibitem[Senge et~al.(2014)Senge, Bösner, Dembczyński, Haasenritter, Hirsch,
  Donner-Banzhoff, and Hüllermeier]{senge_2014_ReliableClassificationLearning}
R.~Senge, S.~Bösner, K.~Dembczyński, J.~Haasenritter, O.~Hirsch,
  N.~Donner-Banzhoff, and E.~Hüllermeier.
\newblock Reliable classification: {{Learning}} classifiers that distinguish
  aleatoric and epistemic uncertainty.
\newblock \emph{Information Sciences}, 255:\penalty0 16--29, 2014.

\bibitem[Shannon(1948)]{shannon_1948_MathematicalTheoryCommunication}
C.~E. Shannon.
\newblock A {{Mathematical Theory}} of {{Communication}}.
\newblock \emph{The Bell System Technical Journal}, 27:\penalty0 379--423,
  1948.

\bibitem[Shelmanov et~al.(2021)Shelmanov, Puzyrev, Kupriyanova, Belyakov,
  Larionov, Khromov, Kozlova, Artemova, Dylov, and
  Panchenko]{shelmanov_2021_ActiveLearningSequence}
A.~Shelmanov, D.~Puzyrev, L.~Kupriyanova, D.~Belyakov, D.~Larionov, N.~Khromov,
  O.~Kozlova, E.~Artemova, D.~V. Dylov, and A.~Panchenko.
\newblock Active {{Learning}} for {{Sequence Tagging}} with {{Deep Pre-trained
  Models}} and {{Bayesian Uncertainty Estimates}}.
\newblock In \emph{Proceedings of the 16th {{Conference}} of the {{EACL}}},
  pages 1698--1712. {Association for Computational Linguistics}, 2021.

\bibitem[Sicking et~al.(2020)Sicking, Akila, Pintz, Wirtz, Fischer, and
  Wrobel]{DBLP:journals/corr/abs-2012-12687}
J.~Sicking, M.~Akila, M.~Pintz, T.~Wirtz, A.~Fischer, and S.~Wrobel.
\newblock Second-moment loss: {A} novel regression objective for improved
  uncertainties.
\newblock \emph{CoRR}, abs/2012.12687, 2020.

\bibitem[Smith and Gal(2018)]{smith_2018_UnderstandingMeasuresUncertaintya}
L.~Smith and Y.~Gal.
\newblock Understanding {{Measures}} of {{Uncertainty}} for {{Adversarial
  Example Detection}}, 2018.

\bibitem[Tan and Le(2019)]{tan_2019_EfficientNetRethinkingModel}
M.~Tan and Q.~V. Le.
\newblock {{EfficientNet}}: {{Rethinking Model Scaling}} for {{Convolutional
  Neural Networks}}.
\newblock In \emph{{{ICML}}}, 2019.

\bibitem[Turkoglu et~al.(2022)Turkoglu, Becker, Gündüz, Rezaei, Bischl,
  Daudt, D'Aronco, Wegner, and
  Schindler]{turkoglu_2022_FiLMEnsembleProbabilisticDeep}
M.~O. Turkoglu, A.~Becker, H.~A. Gündüz, M.~Rezaei, B.~Bischl, R.~C. Daudt,
  S.~D'Aronco, J.~D. Wegner, and K.~Schindler.
\newblock {FiLM-Ensemble}: {{Probabilistic Deep Learning}} via {{Feature-wise
  Linear Modulation}}.
\newblock Accepted at the Thirty-sixth Conference on Neural Information
  Processing Systems (NeurIPS), 2022.

\bibitem[Wilson(2020)]{wilson_2020_CaseBayesianDeep}
A.~G. Wilson.
\newblock The {{Case}} for {{Bayesian Deep Learning}}, 2020.

\bibitem[Winkler et~al.(2022)Winkler, Ojeda, and
  Opper]{winkler_2022_StochasticControlBayesian}
L.~Winkler, C.~Ojeda, and M.~Opper.
\newblock Stochastic {{Control}} for {{Bayesian Neural Network Training}}.
\newblock \emph{Entropy}, 24, 2022.

\bibitem[Woo(2022)]{woo_2022_AnalyticMutualInformation}
J.~O. Woo.
\newblock Analytic {{Mutual Information}} in {{Bayesian Neural Networks}},
  2022.

\bibitem[Wu et~al.(2020)Wu, Knill, Gales, and
  Malinin]{wu_2020_EnsembleApproachesUncertainty}
X.~Wu, K.~M. Knill, M.~J. Gales, and A.~Malinin.
\newblock Ensemble {{Approaches}} for {{Uncertainty}} in {{Spoken Language
  Assessment}}.
\newblock In \emph{Interspeech 2020}, pages 3860--3864. {ISCA}, 2020.

\end{thebibliography}

\newpage
\begin{appendices}
\appendixpage

\section{Experimental Details} \label{app:expdetails}

In the following, we list the most important training configurations used to generate our results.
The full experimental code is hosted in a public repository\footnote{\href{https://github.com/lisa-wm/entropybaseduq}{\texttt{https://github.com/lisa-wm/entropybaseduq}}}.

\paragraph{Software}
Our codebase is written in \texttt{Python}.
It chiefly relies on the \texttt{PyTorch} \citep{paszke_2019_PyTorchImperativeStyle}, \texttt{PyTorch Lightning} \citep{lightningai_2023_PyTorchLightningTrain}, \texttt{Laplace Redux} \citep{daxberger_2021_LaplaceReduxEffortless}, and \texttt{scikit-learn} \citep{scikit-learn} libraries.

\paragraph{Datasets}
The real-world computer vision tasks are \texttt{CIFAR10} \citep{Krizhevsky2009learning} and \texttt{MNIST} \citep{lecun_1998_GradientBasedLearningApplied}.
Both contain ten balanced classes.
We further synthesize rectangles (white-on-black), where the class label is determined by whether height $>$ width or \textit{vice versa}, and random non-convex polygons (white-on-black) with 3--5 vertices.
These datasets comprise 60k (10k) training (test) samples.
The tabular classification problem is created via \texttt{scikit-learn}'s \texttt{make\_classification} function, using two features (and four classes.
Here, we generate 6k (1k) training (test) samples.

\paragraph{Base learners}
Our probabilistic classifiers all combine some base learners into an explicit (deep ensemble, random forest) or implicit (Laplace approximation) ensemble.
We train \texttt{EfficientNet-B7} (approx. 64m parameters; \citet{tan_2019_EfficientNetRethinkingModel}) for \texttt{CIFAR10} and a small convolutional network (three convolutional layers with ReLU activation; approx. 62k parameters) for \texttt{MNIST} and the rectangle/polygon images. 
In the tabular classification problem, we use a random forest with a maximum tree depth of ten as well as single-hidden-layer MLPs with a hidden layer size of ten, adopting the default parameters from \texttt{scikit-learn} unless stated otherwise.
Ensemble size is set to $M = 10$.

\paragraph{Training Configurations}
We use an SGD optimizer (momentum 0.9), a learning rate schedule with cosine annealing, where the initial learning rate is set to $10^{-2}$, and weight decay ($5 \times 10^{-4}$).
Training runs for a maximum of 200 epochs at batch size 256 with early stopping if validation loss does not improve over five consecutive epochs (evaluated on a validation set containing 10\% of the training data).  

\section{Additional Results}

\subsection{Increasing Data Noise} \label{app:labelnoise}

\begin{figure}[H]
    \centering
    \includegraphics[width=0.4\textwidth]{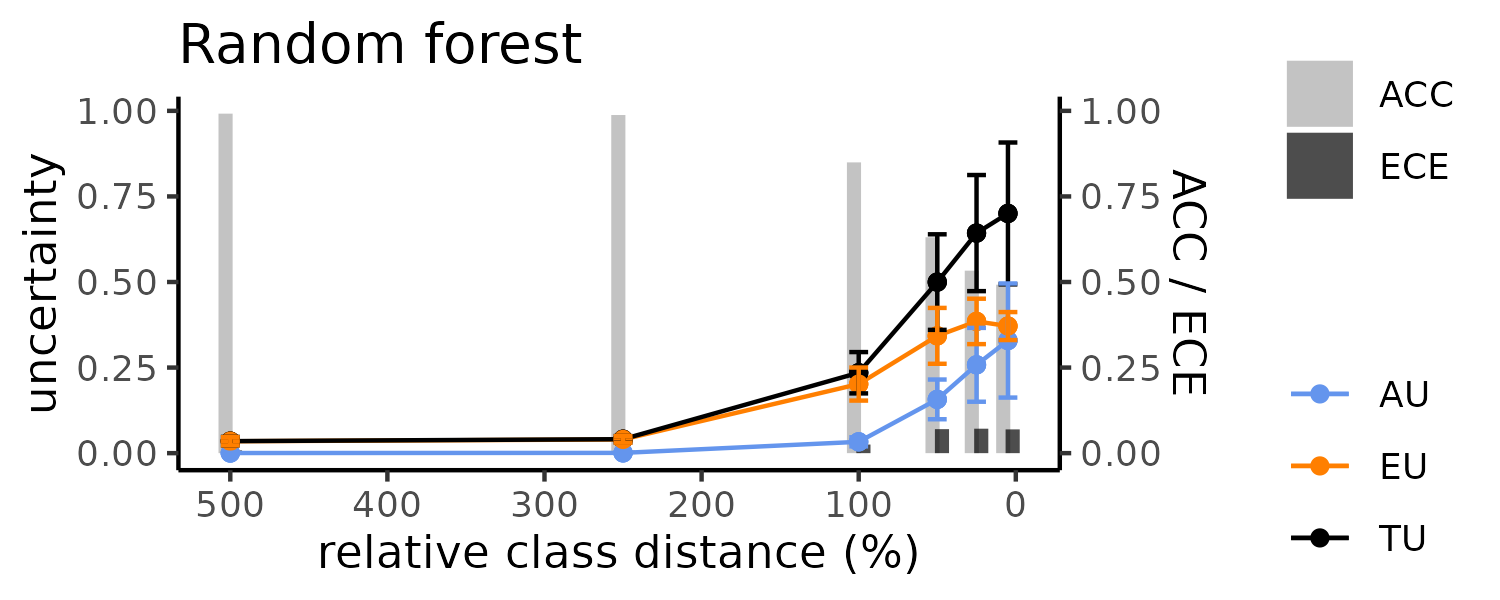}
    \caption{Entropy-based uncertainty for increasing class overlap (tabular data).}
    \label{fig:uvsdistance_rf}
\end{figure}

Compared to the ensemble of MLPs\footnote{
In the tabular classification task, we bootstrap the data for the MLP ensemble to make it directly comparable to the random forest that relies on this technique.
}, the random forest (Fig.~\ref{fig:uvsdistance_rf}) reacts in both uncertainty components when class overlap is increased.

\begin{figure}[H]
    \centering
    \includegraphics[width=0.4\textwidth]{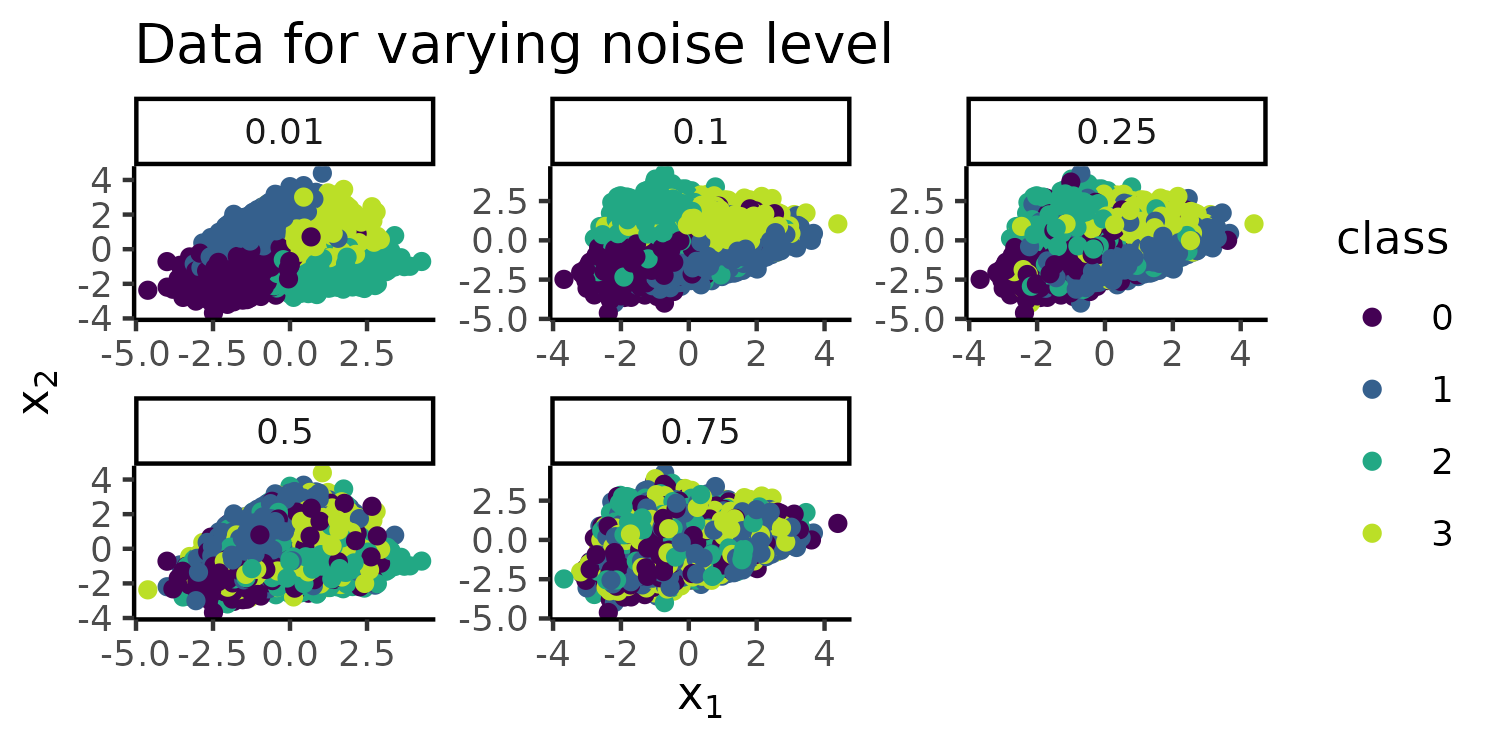}
    \caption{Tabular data with two features and four classes for increasing noise level.}
    \label{fig:uvsnoise_data}
\end{figure}

In order to simulate label noise, we randomly change classes for a varying share (1\%--75\%) of observations in the tabular classification task, leading to datasets as depicted in Fig.~\ref{fig:uvsnoise_data}.

\paragraph{Expected Behavior}
AU picks up with increasing noise level.
Since learner capacity remains fixed, it is reasonable to assume that EU also rises to some extent when the decision boundaries become more complex with mounting degree of dataset contamination.

\begin{figure}[H] \centering
    \includegraphics[width=0.4\textwidth]{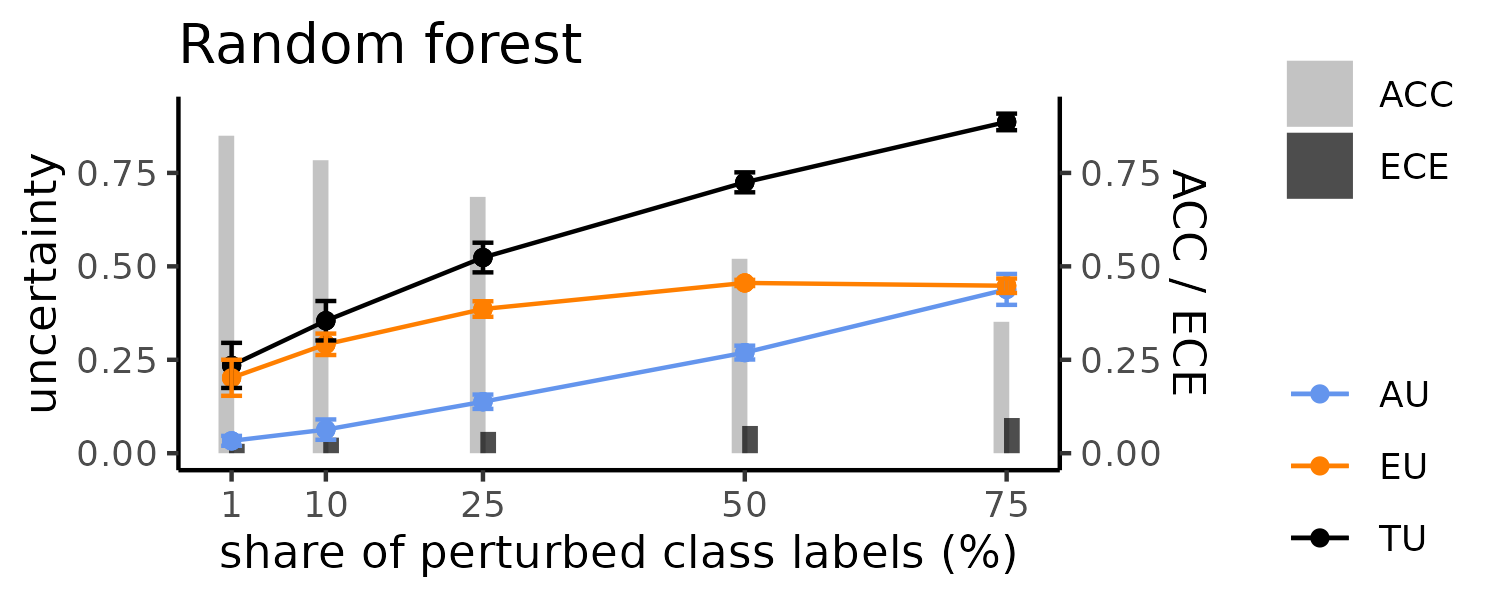}
      \includegraphics[width=0.4\textwidth]{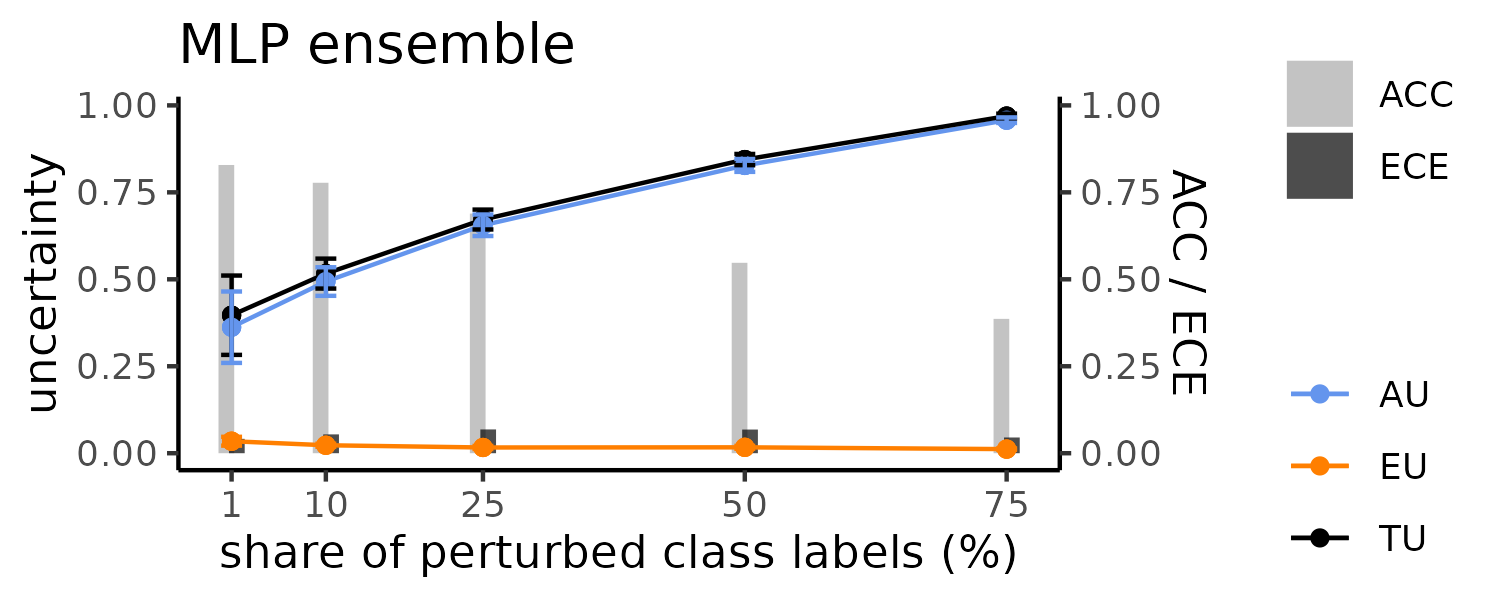}
  \caption{Entropy-based uncertainty for increasing label noise (tabular data).}
  \label{fig:uvsnoise}
\end{figure}

\paragraph{Observed Behavior}
As observed in the experiments modifying image resolution and class overlap, we find that AU duly increases for a rising noise level, though it remains moderate for the random forest even in the most extreme scenario (Fig.~\ref{fig:uvsnoise}), where three out of four labels are assigned randomly.
EU goes up slightly for the random forest, as presumed, but remains ultra-low for every value of the ablation with the MLP ensemble.

\subsection{Number of Ensemble Members} \label{app:nmembers}

We study for the tabular classification problem how different ensemble sizes (2--50) affect the uncertainty estimates.

\paragraph{Expected Behavior}
There should be no systematic pattern except for possible volatility for very small ensemble sizes, where the finite-ensemble estimator might have larger bias.

\begin{figure}[H]
  \centering
      \includegraphics[width=0.4\textwidth]{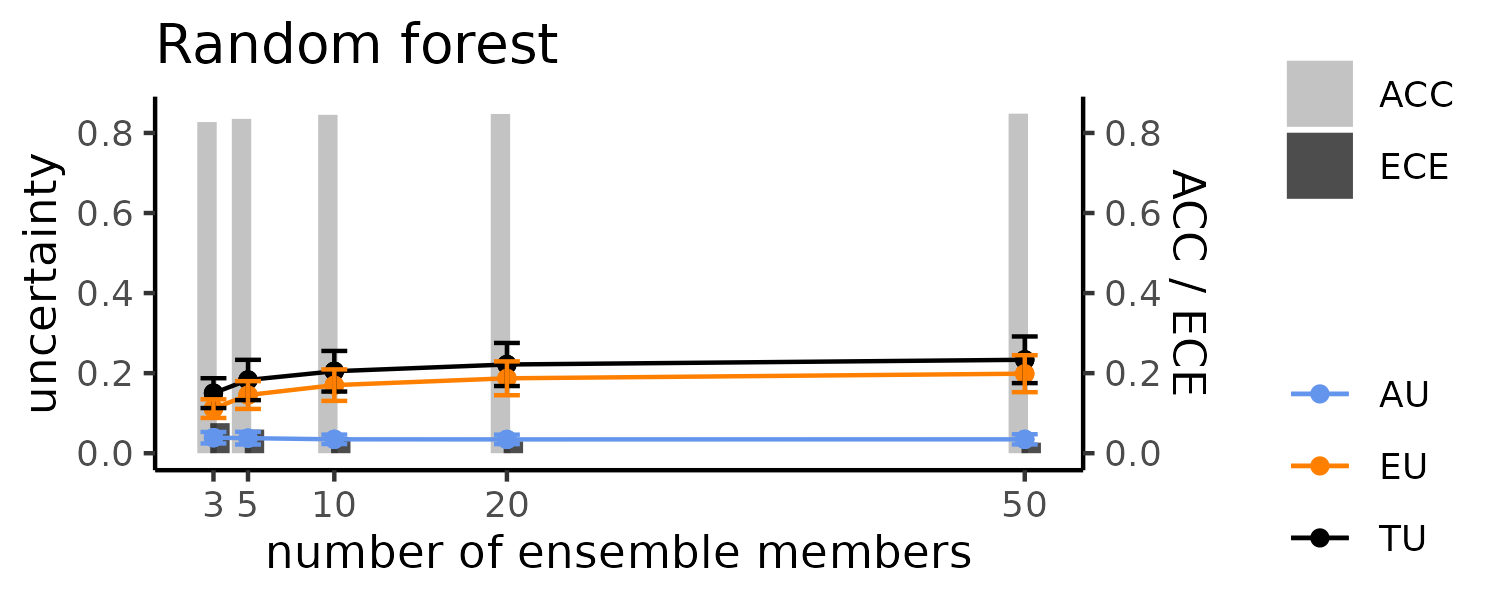}
      \includegraphics[width=0.4\textwidth]{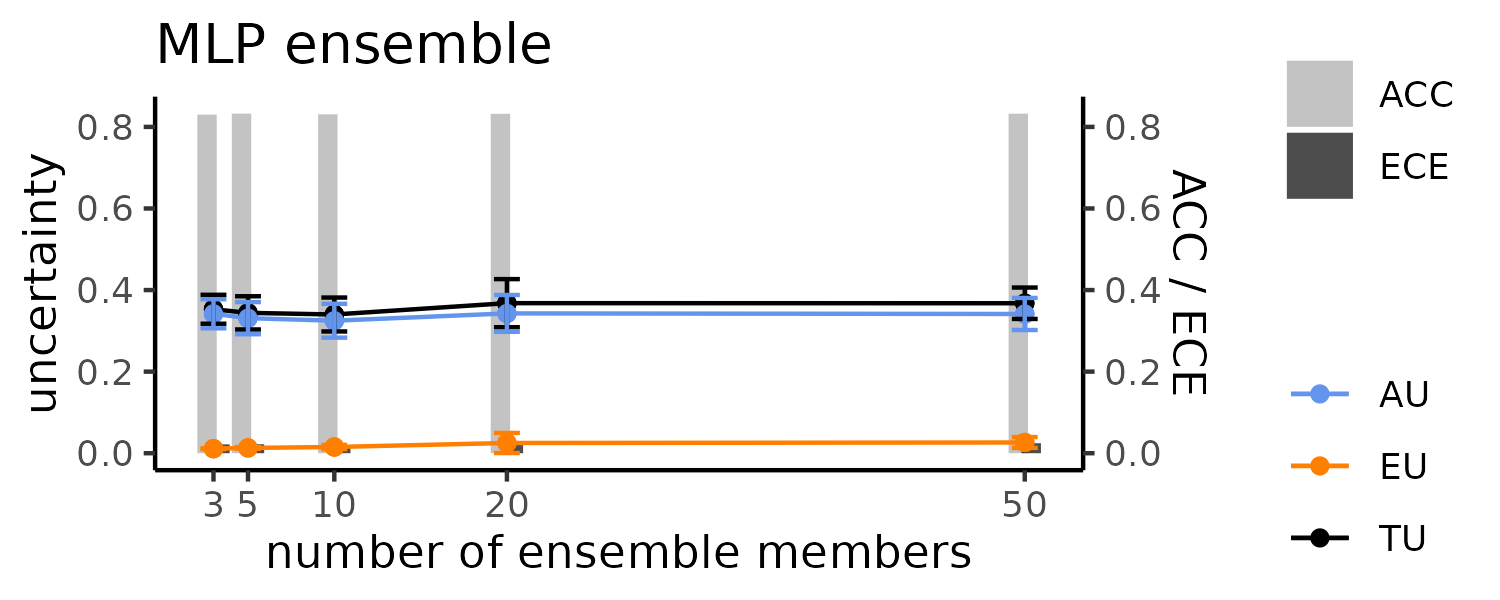}
  \caption{Entropy-based uncertainty for increasing number of ensemble members (tabular data).}
  \label{fig:uvsmembers}
\end{figure}

\paragraph{Observed Behavior}
The results are indeed fairly stable for different values of $M$ (Fig.~\ref{fig:uvsmembers}).
Again, the overall levels of reported uncertainty differ considerably between the learners.

We also compute the uncertainty measures for the computer vision tasks, where such large ensembles are prohibitively expensive, that result from using $M = 5$.
Tables~\ref{tab:samples_la_mnist}--\ref{tab:res_de_cifar} show the uncertainty values as an average over all possible five-member ensembles that can be constructed from the ten original predictions (we can compute this \textit{ex post} since ensemble size does not affect training for either of the used probabilistic learners: deep ensembles are trained in parallel with no shared loss propagation, and Laplace approximation is an inherently \textit{ex-post} approach anyway).
The results are quite robust here as well (with some exceptions for the particularly noisy settings, such as 1\% sample size).

\subsection{Base Learner Complexity} \label{app:complexity}

Lastly, we investigate the effect of changing the base learner's capacity in the random forest and ensemble of MLPs.
As a a proxy for capacity, we use maximum tree depth and hidden-layer size, respectively.

\paragraph{Expected Behavior}
Initially, AU should decrease when base learners get more capacity so they can fit more varied distributions, express their confidence more adequately and achieve better calibration.
Similarly, the additional complexity might result in higher EU because the base learners have more freedom for disagreement.

\begin{figure}[H]
  \centering
      \includegraphics[width=0.4\textwidth]{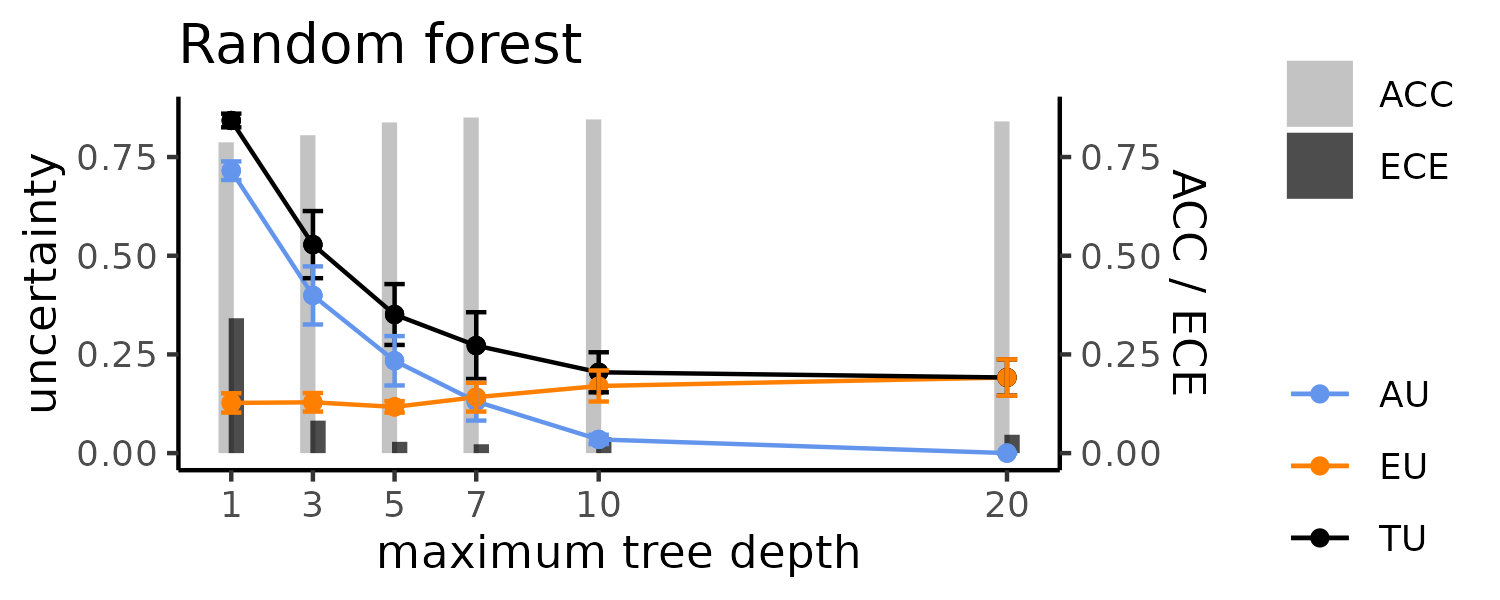}
      \includegraphics[width=0.4\textwidth]{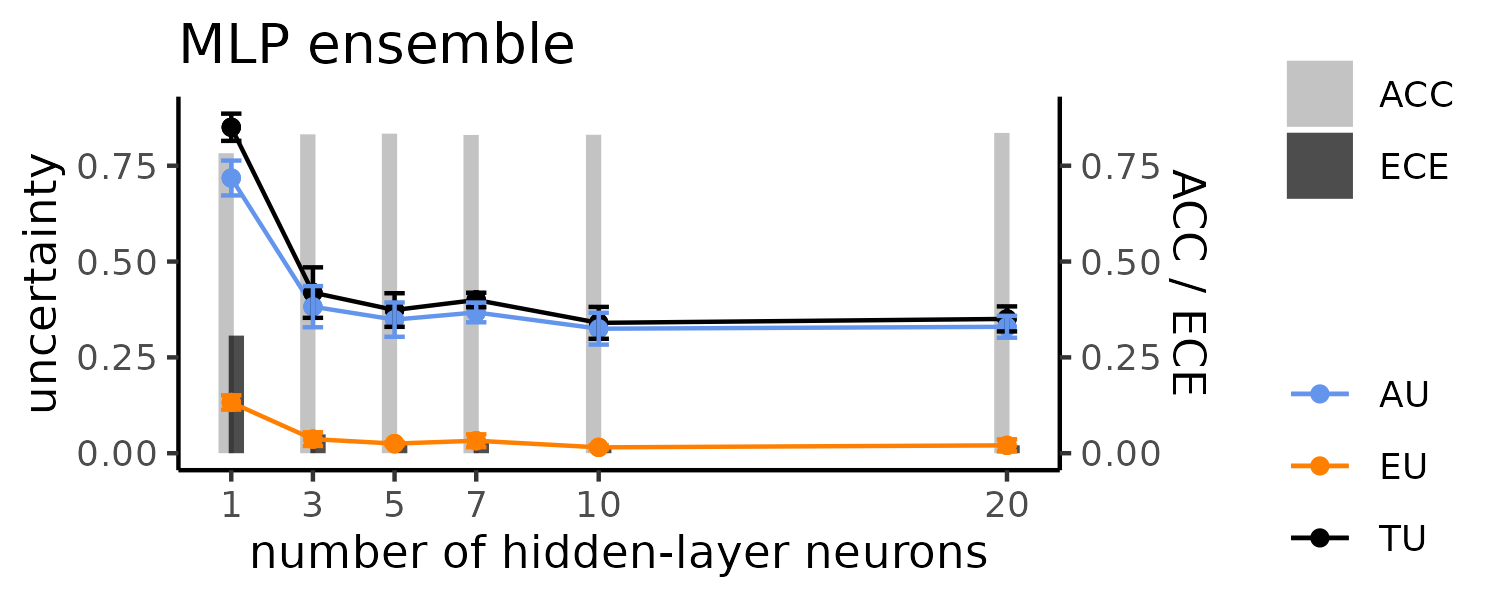}
  \caption{Increasing base learner complexity}
  \label{fig:uvscomplexity}
\end{figure}

\paragraph{Observed Behavior}
We find that AU indeed reduces considerably for more complex base learners (Fig.~\ref{fig:uvscomplexity}), especially for the random forest, which appears to overstate AU when the base learners are very simple (resulting in high calibration error).
The strong effect is quite striking and might be overlooked as performance is relatively stable, again underlining that accuracy, calibration and uncertainty must be considered jointly. 
EU, on the other hand, does not change much -- apparently, relation between capacity and reported AU is quite consistent across base learners and does not provoke more conflict when the ensemble members obtain more freedom.

\onecolumn

\begin{table}
    \centering
        \caption{Results for sample size with ensemble of $M = 5$. Mean and standard deviation are obtained by aggregating over all possible ensembles of size five that can be sampled from the ten predictions of the original experiment.}
    \label{tab:samples_la_mnist}
\begin{tabular}{|l|l|l|l|l|r|r|r|}
\hline
\textbf{Experiment} & \textbf{Case} & \textbf{Probabilistic learner} & \textbf{Dataset} & \textbf{Measure} & \textbf{Mean} & \textbf{Standard deviation} & $M = 10$\\
\hline
sample size & 1 & Laplace approximation & MNIST & TU & 0.5918 & 0.0451 & 0.7754\\
\hline
sample size & 1 & Laplace approximation & MNIST & AU & 0.0091 & 0.0012 & 0.0091\\
\hline
sample size & 1 & Laplace approximation & MNIST & EU & 0.5827 & 0.0449 & 0.7663\\
\hline
sample size & 2 & Laplace approximation & MNIST & TU & 0.5794 & 0.0436 & 0.7419\\
\hline
sample size & 2 & Laplace approximation & MNIST & AU & 0.0386 & 0.0037 & 0.0388\\
\hline
sample size & 2 & Laplace approximation & MNIST & EU & 0.5408 & 0.0416 & 0.7031\\
\hline
sample size & 5 & Laplace approximation & MNIST & TU & 0.5370 & 0.0416 & 0.6716\\
\hline
sample size & 5 & Laplace approximation & MNIST & AU & 0.0631 & 0.0053 & 0.0634\\
\hline
sample size & 5 & Laplace approximation & MNIST & EU & 0.4738 & 0.0391 & 0.6083\\
\hline
sample size & 10 & Laplace approximation & MNIST & TU & 0.4578 & 0.0364 & 0.5760\\
\hline
sample size & 10 & Laplace approximation & MNIST & AU & 0.0447 & 0.0039 & 0.0449\\
\hline
sample size & 10 & Laplace approximation & MNIST & EU & 0.4131 & 0.0341 & 0.5311\\
\hline
sample size & 50 & Laplace approximation & MNIST & TU & 0.1756 & 0.0247 & 0.2147\\
\hline
sample size & 50 & Laplace approximation & MNIST & AU & 0.0354 & 0.0033 & 0.0355\\
\hline
sample size & 50 & Laplace approximation & MNIST & EU & 0.1402 & 0.0221 & 0.1791\\
\hline
sample size & 100 & Laplace approximation & MNIST & TU & 0.0793 & 0.0116 & 0.0947\\
\hline
sample size & 100 & Laplace approximation & MNIST & AU & 0.0223 & 0.0022 & 0.0224\\
\hline
sample size & 100 & Laplace approximation & MNIST & EU & 0.0570 & 0.0096 & 0.0723\\
\hline
\end{tabular}
\end{table}

\begin{table}
    \centering
        \caption{Results for sample size with ensemble of $M = 5$. Mean and standard deviation are obtained by aggregating over all possible ensembles of size five that can be sampled from the ten predictions of the original experiment.}
    \label{tab:samples_de_mnist}
\begin{tabular}{|l|l|l|l|l|r|r|r|}
\hline
\textbf{Experiment} & \textbf{Case} & \textbf{Probabilistic learner} & \textbf{Dataset} & \textbf{Measure} & \textbf{Mean} & \textbf{Standard deviation} & $M = 10$\\
\hline
sample size & 1 & deep ensemble & MNIST & TU & 0.0487 & 0.0110 & 0.0518\\
\hline
sample size & 1 & deep ensemble & MNIST & AU & 0.0343 & 0.0065 & 0.0344\\
\hline
sample size & 1 & deep ensemble & MNIST & EU & 0.0144 & 0.0049 & 0.0174\\
\hline
sample size & 2 & deep ensemble & MNIST & TU & 0.0381 & 0.0042 & 0.0404\\
\hline
sample size & 2 & deep ensemble & MNIST & AU & 0.0257 & 0.0032 & 0.0258\\
\hline
sample size & 2 & deep ensemble & MNIST & EU & 0.0124 & 0.0016 & 0.0146\\
\hline
sample size & 5 & deep ensemble & MNIST & TU & 0.0212 & 0.0017 & 0.0223\\
\hline
sample size & 5 & deep ensemble & MNIST & AU & 0.0152 & 0.0012 & 0.0153\\
\hline
sample size & 5 & deep ensemble & MNIST & EU & 0.0060 & 0.0011 & 0.0070\\
\hline
sample size & 10 & deep ensemble & MNIST & TU & 0.0137 & 0.0011 & 0.0144\\
\hline
sample size & 10 & deep ensemble & MNIST & AU & 0.0098 & 0.0007 & 0.0099\\
\hline
sample size & 10 & deep ensemble & MNIST & EU & 0.0039 & 0.0007 & 0.0045\\
\hline
sample size & 50 & deep ensemble & MNIST & TU & 0.0082 & 0.0007 & 0.0087\\
\hline
sample size & 50 & deep ensemble & MNIST & AU & 0.0051 & 0.0004 & 0.0051\\
\hline
sample size & 50 & deep ensemble & MNIST & EU & 0.0031 & 0.0004 & 0.0036\\
\hline
sample size & 100 & deep ensemble & MNIST & TU & 0.0067 & 0.0008 & 0.0072\\
\hline
sample size & 100 & deep ensemble & MNIST & AU & 0.0042 & 0.0004 & 0.0042\\
\hline
sample size & 100 & deep ensemble & MNIST & EU & 0.0025 & 0.0004 & 0.0030\\
\hline
\end{tabular}
\end{table}

\begin{table}
    \centering
        \caption{Results for sample size with ensemble of $M = 5$. Mean and standard deviation are obtained by aggregating over all possible ensembles of size five that can be sampled from the ten predictions of the original experiment.}
    \label{tab:samples_la_cifar}
\begin{tabular}{|l|l|l|l|l|r|r|r|}
\hline
\textbf{Experiment} & \textbf{Case} & \textbf{Probabilistic learner} & \textbf{Dataset} & \textbf{Measure} & \textbf{Mean} & \textbf{Standard deviation} & $M = 10$\\
\hline
sample size & 1 & Laplace approximation & CIFAR10 & TU & 0.8171 & 0.0629 & 0.8507\\
\hline
sample size & 1 & Laplace approximation & CIFAR10 & AU & 0.6339 & 0.0587 & 0.6364\\
\hline
sample size & 1 & Laplace approximation & CIFAR10 & EU & 0.1832 & 0.0290 & 0.2143\\
\hline
sample size & 2 & Laplace approximation & CIFAR10 & TU & 0.5742 & 0.0368 & 0.6030\\
\hline
sample size & 2 & Laplace approximation & CIFAR10 & AU & 0.4337 & 0.0277 & 0.4354\\
\hline
sample size & 2 & Laplace approximation & CIFAR10 & EU & 0.1405 & 0.0092 & 0.1676\\
\hline
sample size & 5 & Laplace approximation & CIFAR10 & TU & 0.3823 & 0.0247 & 0.4114\\
\hline
sample size & 5 & Laplace approximation & CIFAR10 & AU & 0.2449 & 0.0158 & 0.2459\\
\hline
sample size & 5 & Laplace approximation & CIFAR10 & EU & 0.1374 & 0.0089 & 0.1655\\
\hline
sample size & 10 & Laplace approximation & CIFAR10 & TU & 0.3000 & 0.0199 & 0.3268\\
\hline
sample size & 10 & Laplace approximation & CIFAR10 & AU & 0.1776 & 0.0117 & 0.1783\\
\hline
sample size & 10 & Laplace approximation & CIFAR10 & EU & 0.1224 & 0.0083 & 0.1485\\
\hline
sample size & 50 & Laplace approximation & CIFAR10 & TU & 0.1327 & 0.0089 & 0.1460\\
\hline
sample size & 50 & Laplace approximation & CIFAR10 & AU & 0.0724 & 0.0048 & 0.0727\\
\hline
sample size & 50 & Laplace approximation & CIFAR10 & EU & 0.0603 & 0.0043 & 0.0733\\
\hline
sample size & 100 & Laplace approximation & CIFAR10 & TU & 0.0690 & 0.0047 & 0.0736\\
\hline
sample size & 100 & Laplace approximation & CIFAR10 & AU & 0.0460 & 0.0030 & 0.0461\\
\hline
sample size & 100 & Laplace approximation & CIFAR10 & EU & 0.0231 & 0.0018 & 0.0275\\
\hline
\end{tabular}
\end{table}

\begin{table}
    \centering
        \caption{Results for sample size with ensemble of $M = 5$. Mean and standard deviation are obtained by aggregating over all possible ensembles of size five that can be sampled from the ten predictions of the original experiment.}
    \label{tab:samples_de_cifar}
\begin{tabular}{|l|l|l|l|l|r|r|r|}
\hline
\textbf{Experiment} & \textbf{Case} & \textbf{Probabilistic learner} & \textbf{Dataset} & \textbf{Measure} & \textbf{Mean} & \textbf{Standard deviation} & $M = 10$\\
\hline
sample size & 1 & deep ensemble & CIFAR10 & TU & 0.9022 & 0.0715 & 0.9425\\
\hline
sample size & 1 & deep ensemble & CIFAR10 & AU & 0.7073 & 0.1223 & 0.7100\\
\hline
sample size & 1 & deep ensemble & CIFAR10 & EU & 0.1949 & 0.0770 & 0.2324\\
\hline
sample size & 2 & deep ensemble & CIFAR10 & TU & 0.7189 & 0.0652 & 0.7750\\
\hline
sample size & 2 & deep ensemble & CIFAR10 & AU & 0.4410 & 0.0676 & 0.4428\\
\hline
sample size & 2 & deep ensemble & CIFAR10 & EU & 0.2778 & 0.0404 & 0.3322\\
\hline
sample size & 5 & deep ensemble & CIFAR10 & TU & 0.4204 & 0.0273 & 0.4523\\
\hline
sample size & 5 & deep ensemble & CIFAR10 & AU & 0.2519 & 0.0173 & 0.2529\\
\hline
sample size & 5 & deep ensemble & CIFAR10 & EU & 0.1685 & 0.0113 & 0.1995\\
\hline
sample size & 10 & deep ensemble & CIFAR10 & TU & 0.2826 & 0.0183 & 0.3072\\
\hline
sample size & 10 & deep ensemble & CIFAR10 & AU & 0.1545 & 0.0106 & 0.1551\\
\hline
sample size & 10 & deep ensemble & CIFAR10 & EU & 0.1281 & 0.0082 & 0.1521\\
\hline
sample size & 50 & deep ensemble & CIFAR10 & TU & 0.1318 & 0.0131 & 0.1458\\
\hline
sample size & 50 & deep ensemble & CIFAR10 & AU & 0.0617 & 0.0076 & 0.0620\\
\hline
sample size & 50 & deep ensemble & CIFAR10 & EU & 0.0701 & 0.0057 & 0.0838\\
\hline
sample size & 100 & deep ensemble & CIFAR10 & TU & 0.1064 & 0.0176 & 0.1187\\
\hline
sample size & 100 & deep ensemble & CIFAR10 & AU & 0.0480 & 0.0101 & 0.0482\\
\hline
sample size & 100 & deep ensemble & CIFAR10 & EU & 0.0585 & 0.0078 & 0.0706\\
\hline
\end{tabular}
\end{table}

\begin{table}
    \centering
        \caption{Results for image resolution with ensemble of $M = 5$. Mean and standard deviation are obtained by aggregating over all possible ensembles of size five that can be sampled from the ten predictions of the original experiment.}
    \label{tab:res_la_mnist}
\begin{tabular}{|l|l|l|l|l|r|r|r|}
\hline
\textbf{Experiment} & \textbf{Case} & \textbf{Probabilistic learner} & \textbf{Dataset} & \textbf{Measure} & \textbf{Mean} & \textbf{Standard deviation} & $M = 10$\\
\hline
image resolution & 5 & Laplace approximation & MNIST & TU & 0.7660 & 0.0571 & 0.8540\\
\hline
image resolution & 5 & Laplace approximation & MNIST & AU & 0.3781 & 0.0387 & 0.3796\\
\hline
image resolution & 5 & Laplace approximation & MNIST & EU & 0.3879 & 0.0418 & 0.4744\\
\hline
image resolution & 10 & Laplace approximation & MNIST & TU & 0.6475 & 0.0463 & 0.6996\\
\hline
image resolution & 10 & Laplace approximation & MNIST & AU & 0.3847 & 0.0348 & 0.3862\\
\hline
image resolution & 10 & Laplace approximation & MNIST & EU & 0.2628 & 0.0317 & 0.3134\\
\hline
image resolution & 25 & Laplace approximation & MNIST & TU & 0.1149 & 0.0099 & 0.1261\\
\hline
image resolution & 25 & Laplace approximation & MNIST & AU & 0.0634 & 0.0048 & 0.0636\\
\hline
image resolution & 25 & Laplace approximation & MNIST & EU & 0.0516 & 0.0057 & 0.0624\\
\hline
image resolution & 50 & Laplace approximation & MNIST & TU & 0.0787 & 0.0087 & 0.0923\\
\hline
image resolution & 50 & Laplace approximation & MNIST & AU & 0.0259 & 0.0024 & 0.0260\\
\hline
image resolution & 50 & Laplace approximation & MNIST & EU & 0.0528 & 0.0065 & 0.0663\\
\hline
image resolution & 100 & Laplace approximation & MNIST & TU & 0.0703 & 0.0108 & 0.0833\\
\hline
image resolution & 100 & Laplace approximation & MNIST & AU & 0.0212 & 0.0024 & 0.0213\\
\hline
image resolution & 100 & Laplace approximation & MNIST & EU & 0.0491 & 0.0085 & 0.0620\\
\hline
\end{tabular}
\end{table}

\begin{table}
    \centering
        \caption{Results for image resolution with ensemble of $M = 5$. Mean and standard deviation are obtained by aggregating over all possible ensembles of size five that can be sampled from the ten predictions of the original experiment.}
    \label{tab:res_de_mnist}
\begin{tabular}{|l|l|l|l|l|r|r|r|}
\hline
\textbf{Experiment} & \textbf{Case} & \textbf{Probabilistic learner} & \textbf{Dataset} & \textbf{Measure} & \textbf{Mean} & \textbf{Standard deviation} & $M = 10$\\
\hline
image resolution & 5 & deep ensemble & MNIST & TU & 0.7635 & 0.0487 & 0.7672\\
\hline
image resolution & 5 & deep ensemble & MNIST & AU & 0.7585 & 0.0484 & 0.7615\\
\hline
image resolution & 5 & deep ensemble & MNIST & EU & 0.0050 & 0.0008 & 0.0056\\
\hline
image resolution & 10 & deep ensemble & MNIST & TU & 0.5378 & 0.0350 & 0.5413\\
\hline
image resolution & 10 & deep ensemble & MNIST & AU & 0.5272 & 0.0344 & 0.5292\\
\hline
image resolution & 10 & deep ensemble & MNIST & EU & 0.0107 & 0.0013 & 0.0121\\
\hline
image resolution & 25 & deep ensemble & MNIST & TU & 0.0519 & 0.0039 & 0.0537\\
\hline
image resolution & 25 & deep ensemble & MNIST & AU & 0.0421 & 0.0030 & 0.0423\\
\hline
image resolution & 25 & deep ensemble & MNIST & EU & 0.0098 & 0.0012 & 0.0114\\
\hline
image resolution & 50 & deep ensemble & MNIST & TU & 0.0088 & 0.0008 & 0.0093\\
\hline
image resolution & 50 & deep ensemble & MNIST & AU & 0.0058 & 0.0004 & 0.0059\\
\hline
image resolution & 50 & deep ensemble & MNIST & EU & 0.0030 & 0.0004 & 0.0035\\
\hline
image resolution & 100 & deep ensemble & MNIST & TU & 0.0074 & 0.0006 & 0.0079\\
\hline
image resolution & 100 & deep ensemble & MNIST & AU & 0.0046 & 0.0004 & 0.0046\\
\hline
image resolution & 100 & deep ensemble & MNIST & EU & 0.0028 & 0.0003 & 0.0033\\
\hline
\end{tabular}
\end{table}

\begin{table}
    \centering
        \caption{Results for image resolution with ensemble of $M = 5$. Mean and standard deviation are obtained by aggregating over all possible ensembles of size five that can be sampled from the ten predictions of the original experiment.}
    \label{tab:res_la_cifar}
\begin{tabular}{|l|l|l|l|l|r|r|r|}
\hline
\textbf{Experiment} & \textbf{Case} & \textbf{Probabilistic learner} & \textbf{Dataset} & \textbf{Measure} & \textbf{Mean} & \textbf{Standard deviation} & $M = 10$\\
\hline
image resolution & 5 & Laplace approximation & CIFAR10 & TU & 0.7137 & 0.0451 & 0.7171\\
\hline
image resolution & 5 & Laplace approximation & CIFAR10 & AU & 0.7093 & 0.0448 & 0.7122\\
\hline
image resolution & 5 & Laplace approximation & CIFAR10 & EU & 0.0043 & 0.0003 & 0.0049\\
\hline
image resolution & 10 & Laplace approximation & CIFAR10 & TU & 0.2676 & 0.0169 & 0.2697\\
\hline
image resolution & 10 & Laplace approximation & CIFAR10 & AU & 0.2593 & 0.0164 & 0.2603\\
\hline
image resolution & 10 & Laplace approximation & CIFAR10 & EU & 0.0083 & 0.0005 & 0.0095\\
\hline
image resolution & 25 & Laplace approximation & CIFAR10 & TU & 0.1151 & 0.0073 & 0.1176\\
\hline
image resolution & 25 & Laplace approximation & CIFAR10 & AU & 0.1007 & 0.0064 & 0.1011\\
\hline
image resolution & 25 & Laplace approximation & CIFAR10 & EU & 0.0144 & 0.0010 & 0.0165\\
\hline
image resolution & 50 & Laplace approximation & CIFAR10 & TU & 0.0835 & 0.0055 & 0.0890\\
\hline
image resolution & 50 & Laplace approximation & CIFAR10 & AU & 0.0545 & 0.0036 & 0.0547\\
\hline
image resolution & 50 & Laplace approximation & CIFAR10 & EU & 0.0290 & 0.0021 & 0.0343\\
\hline
image resolution & 100 & Laplace approximation & CIFAR10 & TU & 0.0690 & 0.0047 & 0.0736\\
\hline
image resolution & 100 & Laplace approximation & CIFAR10 & AU & 0.0460 & 0.0030 & 0.0461\\
\hline
image resolution & 100 & Laplace approximation & CIFAR10 & EU & 0.0231 & 0.0018 & 0.0275\\
\hline
\end{tabular}
\end{table}

\begin{table}
    \centering
        \caption{Results for image resolution with ensemble of $M = 5$. Mean and standard deviation are obtained by aggregating over all possible ensembles of size five that can be sampled from the ten predictions of the original experiment.}
    \label{tab:res_de_cifar}
\begin{tabular}{|l|l|l|l|l|r|r|r|}
\hline
\textbf{Experiment} & \textbf{Case} & \textbf{Probabilistic learner} & \textbf{Dataset} & \textbf{Measure} & \textbf{Mean} & \textbf{Standard deviation} & $M = 10$\\
\hline
image resolution & 5 & deep ensemble & CIFAR10 & TU & 0.7351 & 0.0467 & 0.7425\\
\hline
image resolution & 5 & deep ensemble & CIFAR10 & AU & 0.7012 & 0.0446 & 0.7040\\
\hline
image resolution & 5 & deep ensemble & CIFAR10 & EU & 0.0339 & 0.0027 & 0.0386\\
\hline
image resolution & 10 & deep ensemble & CIFAR10 & TU & 0.3727 & 0.0248 & 0.3900\\
\hline
image resolution & 10 & deep ensemble & CIFAR10 & AU & 0.2752 & 0.0197 & 0.2763\\
\hline
image resolution & 10 & deep ensemble & CIFAR10 & EU & 0.0975 & 0.0067 & 0.1137\\
\hline
image resolution & 25 & deep ensemble & CIFAR10 & TU & 0.2084 & 0.0187 & 0.2265\\
\hline
image resolution & 25 & deep ensemble & CIFAR10 & AU & 0.1134 & 0.0126 & 0.1138\\
\hline
image resolution & 25 & deep ensemble & CIFAR10 & EU & 0.0950 & 0.0068 & 0.1127\\
\hline
image resolution & 50 & deep ensemble & CIFAR10 & TU & 0.1174 & 0.0088 & 0.1302\\
\hline
image resolution & 50 & deep ensemble & CIFAR10 & AU & 0.0527 & 0.0045 & 0.0529\\
\hline
image resolution & 50 & deep ensemble & CIFAR10 & EU & 0.0648 & 0.0046 & 0.0773\\
\hline
image resolution & 100 & deep ensemble & CIFAR10 & TU & 0.1045 & 0.0137 & 0.1160\\
\hline
image resolution & 100 & deep ensemble & CIFAR10 & AU & 0.0480 & 0.0084 & 0.0482\\
\hline
image resolution & 100 & deep ensemble & CIFAR10 & EU & 0.0565 & 0.0055 & 0.0679\\
\hline
\end{tabular}
\end{table}

\end{appendices}

\end{document}